\documentclass{article}
\usepackage[preprint]{neurips_2025}

\usepackage[utf8]{inputenc} % allow utf-8 input
\usepackage[T1]{fontenc}    % use 8-bit T1 fonts
\usepackage{hyperref}       % hyperlinks
\usepackage{url}            % simple URL typesetting
\usepackage{booktabs}       % professional-quality tables
\usepackage{enumitem}
\usepackage[table]{xcolor}
\usepackage{pgfplotstable}
\usepackage{amsfonts}       % blackboard math symbols
\usepackage{makecell}
\usepackage{graphicx}
\usepackage{pifont}   % for \ding{55}
\usepackage{amssymb}
\usepackage[table]{xcolor}
\usepackage{booktabs}
\usepackage{svg}
\usepackage{amsmath}
\usepackage{pdflscape}
\usepackage{nicefrac}       % compact symbols for 1/2, etc.
\usepackage{microtype}      % microtypography
\usepackage{xcolor}         % colors
\usepackage{graphicx}       % graphs
\usepackage{multicol}       % two-column layout
\usepackage{mdframed}
\usepackage{multirow}
\usepackage{CJKutf8}
\usepackage{amsmath}
\usepackage{xspace}
\usepackage{graphicx,subcaption} 
\usepackage[most]{tcolorbox}
\usepackage{wrapfig}
\usepackage{caption}    % in your preamble
\usepackage{tcolorbox}
\tcbuselibrary{breakable}
\usepackage[table]{xcolor}           % for \cellcolor and friends
\usepackage{booktabs}                % for \toprule, \midrule, \bottomrule
\usepackage{pgfplots}                % core pgfplots
\usepackage{pgfplotstable}           % core table routines

\def\ourmetric{\texttt{SR}\xspace}

\tcbset{
  palebluebox/.style={
    enhanced jigsaw,     % use the jigsaw engine for flexible breaking
    breakable,           % allow the box to split across pages
    colback=gray!10!white,
    colframe=gray!30!white,
    boxrule=0pt,
    sharp corners,
    width=\textwidth,
    before skip=10pt,
    after skip=10pt,
    left=8pt, right=8pt, top=3pt, bottom=3pt,
  }
}

\title{\includegraphics[width=0.08\textwidth]{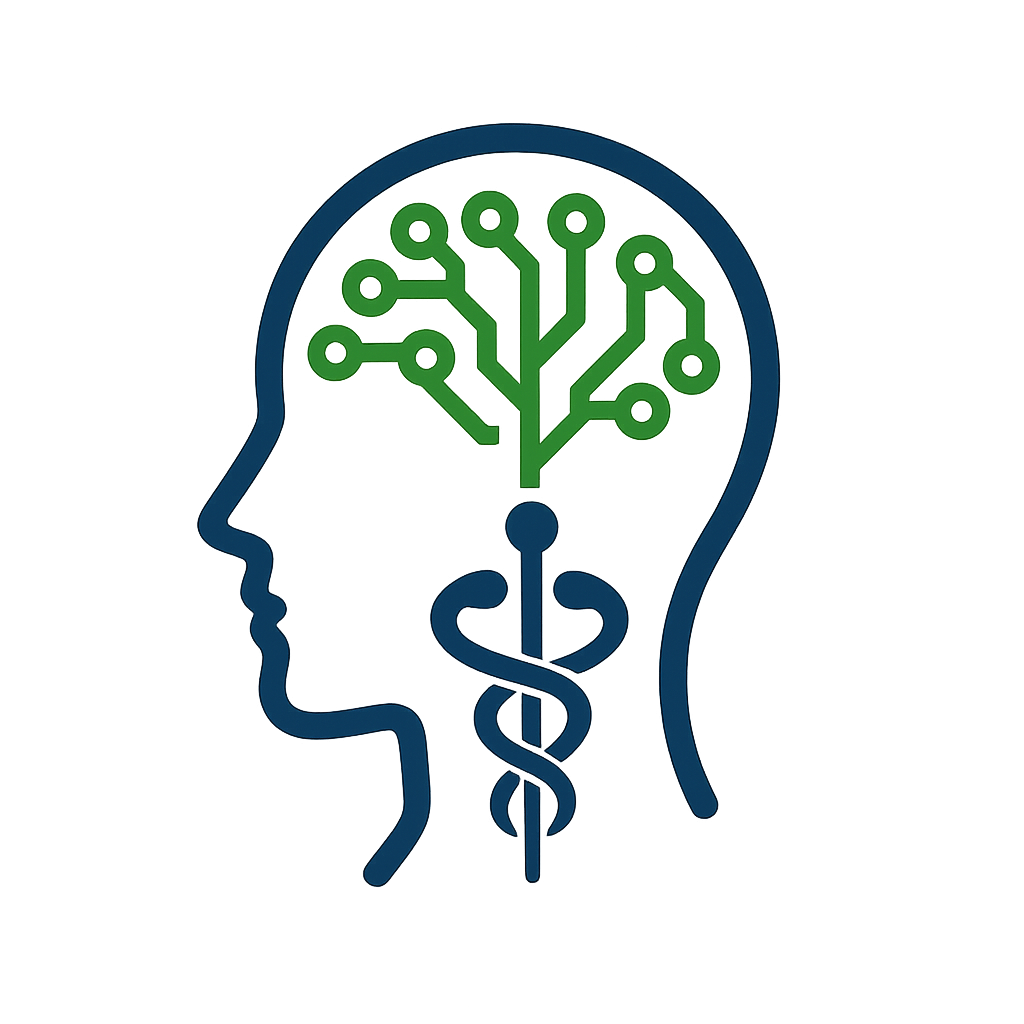}
  MedPAIR: Measuring \underline{P}hysicians and \underline{AI} \underline{R}elevance Alignment in Medical Question Answering}
\author{
Yuexing Hao$^{1,2}$, 
Kumail Alhamoud$^1$, 
Hyewon Jeong$^1$,
Haoran Zhang$^1$, 
Isha Puri$^1$,\\ 
\textbf{Philip Torr$^3$, Mike Schaekermann$^6$, Ariel D. Stern$^{4, 5}$, Marzyeh Ghassemi$^{1}$\thanks{Corresponding author: mghassem@mit.edu}}\\
\\
$^1$MIT, $^2$Cornell University, $^3$Oxford University,\\ $^4$Hasso Plattner Institute, $^5$University of Potsdam $^6$Independent Research
}

\begin{document}
\maketitle

\begin{abstract}

Large Language Models (LLMs) have demonstrated remarkable performance on various medical question-answering (QA) benchmarks, including standardized medical exams. However, correct answers alone do not ensure correct logic, and models may reach accurate conclusions through flawed processes. In this study, we introduce the \textbf{MedPAIR} (\underline{Med}ical Dataset Comparing \underline{P}hysicians and \underline{AI} \underline{R}elevance Estimation and Question Answering) dataset to evaluate how physician trainees and LLMs prioritize relevant information when answering QA questions. We obtain annotations on 1,300 QA pairs from 36 physician trainees, labeling each sentence within the question components for relevance. We compare these relevance estimates to those for LLMs, and further evaluate the impact of these ``relevant'' subsets on downstream task performance for both physician trainees and LLMs. We find that LLMs are frequently not aligned with the content relevance estimates of physician trainees. After filtering out physician trainee-labeled irrelevant sentences, accuracy improves for both the trainees and the LLMs. All LLM and physician trainee-labeled data are available at: \url{http://medpair.csail.mit.edu/}.
\end{abstract}
% https://github.com/YuexingHao/MedPAIR

% Large Language Models (LLMs) have demonstrated remarkable performance on various medical question-answering benchmarks, including standardized exams. However, correct answers alone do not guarantee correct reasoning, as models can arrive at accurate conclusions through flawed processes. In this study, we introduce the comparative reasoning benchmark MIND-MAP (Medical Inference and Nuanced Decision-Making between AI and Physicians) to evaluate how clinicians and LLMs prioritize information before answering clinical questions. MIND-MAP comprises clinician annotations on 200 patient cases, classifying question components as highly relevant, marginally relevant, or irrelevant to the correct answer. We also implemented a structured intervention in which clinicians guided LLMs to focus on relevant information. Our results reveal frequent misalignment between clinician and LLM reasoning and underscore the limitations of current model behavior as well as the importance of aligning AI reasoning with clinical standards.

% \keywords{Reasoning, Large Language Models, Medical Question-Answering (QA)}

\section{Introduction}
Large language models (LLMs) have shown strong performance across a range of medical tasks, with systems like GPT-4 and MedPaLM outperforming human averages on standardized medical examinations \cite{cabral_clinical_2024, mcduff_towards_2025}. However, many tasks do not reflect the complexity of real-world use cases \cite{xia_cares_2024}, and high performance on exam-style datasets may overstate a LLM’s generalizability~\cite{li_mediq_2024}. In human-facing settings, it is crucial to understand how models filter and prioritize relevant information \cite{park_does_2015}. %To date, there is no dataset comparing the similarity of LMMs and human expert prioritization of data, or how this may impact task performance.

Estimation of contextual relevance is a critical aspect in many applications. Techniques such as semantic entropy \cite{farquhar_detecting_2024}, influence functions \cite{choe_what_2024}, context attribution \cite{cohen-wang_learning_2025, liu_attribot_2024}, and evidence inference \cite{deyoung_eraser_2020} have been employed to assess which elements within a context hold the most importance. Despite these efforts, existing relevance estimations are often imprecise and noisy, with models sometimes producing misleading or overly confident assessments that deviate from human judgment \cite{es_ragas_2024}. Even when estimations appear less noisy, model-generated relevance labels may not concord with those of human experts. This gap is particularly concerning in human-facing domains where alignment with expert judgment is necessary \cite{zhao_moverscore_2019, schuler_context_2025}. 

We focus on the question-answering (QA) in clinical contexts, which reflect how physicians synthesize patient data to address specific concerns. Existing medical QA datasets have driven progress in evaluating LLM's performance on clinically relevant tasks \cite{pal_medmcqa_2022, zhang_survey_2024, jin_rjua-meddqa_2024, soni_radqa_2022}. However, QA benchmarks and leader boards primarily assess final answers, providing limited visibility into the underlying rationale \cite{toma_clinical_2023, adlakha_evaluating_2024}. %To improve our understanding of what context in the QA that LLM used for answering, we first need a comparison study between domain experts and LLM's relevance labels and measure its alignment. 

% Recent work has argued that the absence of expert-annotated reasoning data limits the development of reliable models  \cite{zhou_explainable_2025, tang_medagents_2024}. 

% While recent work in ``chain of thought'' (CoT) reasoning models has provided some potential relevance estimates \cite{lightman_lets_2023}, CoT has practical issues that limit its use. For instance, error compounding in long reasoning chains: a small mistake in an early step (e.g. a slightly wrong lab value interpretation) can cascade through subsequent steps, leading to an incorrect final diagnosis. CoT and other method may also not be accurate reports of what a model is truly using \cite{chuang_lookback_2024}. 

In this work, we curate a \underline{Med}ical Dataset comparing \underline{P}hysician trainees and \underline{AI} \underline{R}elevance estimation and question answering  - \emph{MedPAIR}. We design \emph{MedPAIR} to understand how physician trainees and LLMs select relevant information in structured QA. 
%\emph{MedPAIR} is a first benchmark step to matching the relevance annotated by clinical professional labelers, to that estimated by LLMs. 
%The motivation for \emph{MedPAIR} is to ensure that \textit{what the LLM finds relevant} in a clinical case closely matches \textit{what a physician trainee finds relevant}. 
We collect sentence-level relevance labels on 2000 samples from the four QA benchmark datasets from 36 physician trainees. In parallel, we prompt LLMs to self-report sentence-level relevance~\cite{choi_identifying_2024} and apply ContextCite \cite{cohen-wang_contextcite_2024}, a context attribution framework that maps model outputs to the input sentences most responsible for their generation. This approach allows us to quantify the degree of alignment between human and model assessments of contextual importance. Using these annotations, we evaluate how sentence-level relevance, estimated by either LLMs or humans, affects downstream QA performance. We release the \textbf{first benchmark and open-source dataset of physician trainee-annotated relevance for patient case QA tasks}, enabling direct comparison with LLM-assigned scores.  Our full workflow can be found in Figure \ref{fig:study_design}.

\section{Related Work}

\subsection{Aligning Human and LLM Estimation}

Previous work has shown that limiting input to relevant information can reduce distraction, streamline evidence integration, and reduce memory requirements \cite{khattab_baleen_2021, chuang_selfcite_2025, lu_learn_2022, bucinca_trust_2021}. Ensuring Artificial Intelligence (AI) systems focus on the same input information that physician trainees identify as relevant is crucial to evaluating which clinical details informed each prediction \cite{strong_chatbot_2023, katz_gpt_2024}. This alignment allows physicians to judge the reliability and explainability of AI suggestions and reduces the risk of mistakes caused by extraneous or misinterpreted inputs \cite{chiang_can_2023}. Demonstrating alignment between LLM-selected input context and expert judgment is increasingly recognized as fundamental to earning physician trust in diagnostic AI tools \cite{zhou_explainable_2025, wu_medreason_2025, bansal_does_2021}.  

Effective clinical decision-making often relies on understanding nuanced input information that can reveal critical insights. In a systematic review, Schuler and colleagues identify 946 distinct contextual factors that influence clinical decisions, demonstrating the complexity of integrating these elements into evidence-based reasoning \cite{schuler_context_2025}. For AI systems to be trusted in clinical settings, they must reflect this contextual understanding, prioritizing information in a way that resonates with clinical judgment \cite{gaube_non-task_2023, hao_retrospective_2025, yang_harnessing_2023}. Transparent alignment between AI reasoning and clinician perspectives can reduce the risks of misleading correlations and enhance trust in AI-based clinical decision support \cite{chanda_dermatologist-like_2024, rong_towards_2024}, where alignment improved confidence in AI-assisted diagnoses. Aligning AI models with physicians' nuanced contextual understanding is essential for their acceptance by the medical establishment and effective integration into clinical practice.

% \emph{MedPAIR} offers a apple-to-apple comparison benchmark on the nuanced sentence-level to understand how both physician trainees and LLMs identify input context. This alignment provides critical insight into whether the model focuses on the same clinically relevant details that clinicians consider essential.

\subsection{Challenges in Comparing LLM and Human Relevance Judgments}

Recent work has questioned the reliability of LLMs in consistently judging the relevance of information \cite{chen_reasoning_nodate}. Although models such as GPT-4 demonstrate high average performance across benchmark datasets, studies have documented significant variance in self-reported labels when identical prompts are issued multiple times \cite{cohen-wang_contextcite_2024, gao_enabling_2023, funkquist_citebench_2023}. This inconsistency is attributed to several factors: prompt sensitivity \cite{wang_prompt_2024, razavi_benchmarking_2025, savage_diagnostic_2024}, stochastic decoding procedures, architectural idiosyncrasies of the model, and ambiguity in input data. Even in deterministic settings (e.g., temperature zero), LLMs can produce divergent responses due to underlying randomness or unstable decision boundaries \cite{ahn_prompt-reverse_2025}. Empirical evaluations confirm that model agreement across repeated prompts is rarely perfect, with accuracy fluctuations of up to 10\% depending on task complexity and phrasing \cite{atil_llm_2024}. Further highlighting this gap, recent studies report that between 50\% and 90\% of LLM-generated medical answers are not fully supported by the cited references \cite{wu_automated_2025}. 
There are multiple ways to evaluate LLM's relevance judgments in the input context, for example semantic entropy using probabilistic approaches to detect hallucination
\cite{farquhar_detecting_2024}.

These challenges are particularly pronounced in clinical contexts; widely used medical QA benchmarks provide limited visibility into \textit{how} models interpret context. For example, PubMedQA \cite{jin_pubmedqa_2019} does not offer detailed annotations identifying which parts of the text are crucial for answering the question \cite{singhal_toward_2025}. Similarly, MedQA \cite{jin_what_2021} lacks expert-provided rationales or sentence-level relevance labels. Other datasets, including MedMCQA \cite{pal_medmcqa_2022}, MMLU’s medical subsets \cite{hendrycks_measuring_2021}, MetaMedQA \cite{griot_large_2025}, and MEDIQ \cite{li_mediq_2024}, also prioritize answer correctness.

\section{The MedPAIR Dataset}
\subsection{QA Dataset Setup}

To examine the alignment between physician trainees’ and LLMs’ assessments of relevance, we deliberately concentrate on existing QA pairs grounded in specific patient case scenarios, rather than general evidence-based questions. This focus is intended to better simulate authentic clinical contexts. Therefore,  we draw on our four datasets \textit{Massive Multitask Language Understanding}  (MMLU)-precision medicine (272 QAs) \cite{hendrycks_measuring_2021}, \textit{Medbullets} (298 QAs), \textit{JAMA Clinical Challenge dataset} (1,034 QAs) \cite{chen_benchmarking_2025}, and \textit{MedXpertQA} (2,450 QAs) \cite{zuo_medxpertqa_2025}. The characteristics for each dataset are presented in Appendix section \ref{dataset_explanation}. Each source provides multi-sentence patient case descriptions paired with questions (4-option or 10-option multiple-choice) and answers, offering a rich context for relevance annotation. Such patient vignettes are broadly recognized in the medical and social sciences, including health economics, and physician responses to clinical vignettes have been shown to predict realized billing behavior in the U.S. Medicare system \cite{cutler_physician_2019}.

% --- Figure 1: Study Design ---
\begin{figure}[h!]
  \centering
  \includegraphics[width=\textwidth]{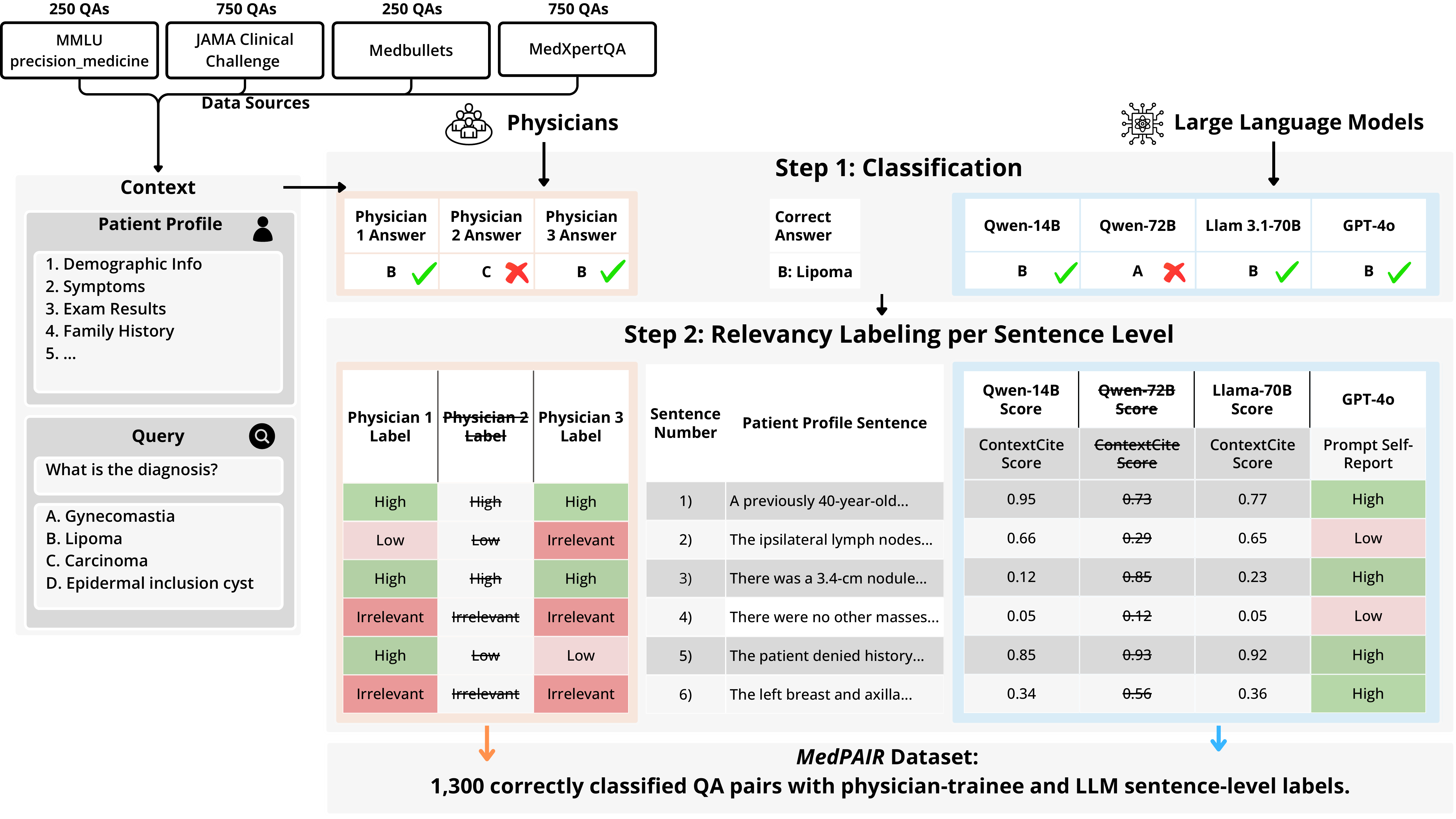}
  \caption{\textbf{Study Design.} We consolidated four QA data sources into two main components: the patient profile and the query. In the first step, 36 physician trainees and 4 LLMs independently selected the most appropriate answer. In the second step, physician trainees annotated the relevance of each sentence within the patient profile, excluding annotations linked to incorrect answers. Majority voting was used to produce binary relevance labels for physician trainees. Concurrently, we employed ContextCite with open-source LLMs (Qwen-14B, Qwen-72B, Llama-70B) to generate relevance scores, while GPT-4o was prompted to replicate the physician annotation process for each sentence following the same instructions.}
  \label{fig:study_design}
\end{figure}

To ensure diversity in patient scenarios, we include only those QAs in which the patient profile contained more than two sentences. Each QA presents a detailed vignette, which is often a long case description covering patient demographic information, symptoms, exam findings, family history, etc. From a combined pool of 4,052 QA pairs, we constructed the final dataset by randomly sampling 250 pairs each from MMLU and Medbullets, and 750 pairs each from JAMA and MedXpertQA, resulting in a curated dataset of 2,000 QA pairs. By pooling these sources, \emph{MedPAIR} covers a wide spectrum of clinical topics and difficulty levels, ensuring that the evaluation is robust across various scenarios from routine to rare conditions. Figure \ref{fig:study_design} shows the \emph{MedPAIR} data curation process.

\subsubsection{Expert Data Annotation}

We partnered with Centaur Labs\footnote{https://centaur.ai} to employ physician trainees (medical students or higher qualifications) annotate the QA pairs. Physician trainees were chosen for their familiarity with medical exam preparation, as these questions are primarily designed for medical students. The demographic information is presented in the appendix table \ref{demographic}.

A total of 36 physician trainees participated (mean age: 26.4), with 77.3\% at the advanced training level and 81.8\% of the labelers had passed the United States Medical Licensing Examination Step 1 Exam. Notably, half of them reported familiarity with using LLMs in clinical contexts, such as integrating tools like ChatGPT into clinical queries or workflows. On average, each labeler spent an average of 3.28 minutes (SD 3.41 min) per QA. When participants arrived at the correct answer and provided sentence-level labeling, they spent on average 3.26 minutes (SD 3.18 min); for incorrect answers with labeling, the mean time was 3.30 minutes (SD 3.62 min). For each case, physician trainees first selected the most appropriate answer and then annotated the relevance of every sentence in the patient profile. A sample physician trainees' annotation is presented in Figure \ref{fig:study_design} (orange boxes). Full study instructions and the pre- and post-survey instruments are provided in the supplementary material. Data collection occurred between March 6 and May 5, 2025.

% To avoid spurious correlations from conditioning relevance on known answers, \emph{MedPAIR} collects relevance judgments from physician trainees without disclosing the correct answer and evaluates model performance on held‐out cases, ensuring that any gains reflect genuine improvements in focus and efficiency rather than answer leakage.

Each QA is annotated by at least three physician trainees and verified to contain at least one correct answer. For each sentence, annotators applied one of three labels: (1) \textbf{High Relevance:} Information that is critical and must be considered to answer the question correctly. These are the key clinical clues or data points that strongly point toward the correct diagnosis or decision. (2) \textbf{Low Relevance:} Information that provides some context or minor clues but is not essential. These details might help rule out alternatives, yet the question could still be answered correctly without them. (3) \textbf{Irrelevant:} Information that is not pertinent to determining the correct answer. These can be distractors or background details included in the vignette that do not impact the outcome in the given context.

This annotation process presents a fine-grained ground truth of relevance for every QA: a trinary label for each sentence in the context, representing the physician trainee consensus on whether that piece of information is pertinent to the question. While obtaining these annotations demanded expert effort, they serve as a gold standard for capturing what physicians deem significant. This level of detailed expert labeling is largely absent from existing medical QA benchmarks, which typically include only the question and answer, without explicit identification of supporting case details and their degree of relevance \cite{chen_benchmarking_2025}.

\subsubsection{LLM Data Annotations}

To directly compare physician trainees’ majority-vote annotations with LLM-generated labels for each QA pair, we annotated the LLM outputs using both ContextCite and a self-reporting prompt. ContextCite scores approximate the model’s attention distribution across sentences \cite{bohnet_attributed_2023}, while self-reported labels capture the model’s own assessment of sentence relevance via prompting. 

Then we performed a sentence-level analysis of their respective annotations to examine this divergence at the sentence level. We examined one closed-source model GPT-4o \cite{openai_gpt-4o_2024} and three open-source models: Qwen-14B, Llama 3.1 Instruct 70B \cite{grattafiori_llama_2024}, and Qwen 2.5-72B \cite{qwen_qwen25_2025}. For GPT-4o, we structured the study the same as physician trainee labeling protocol: we fed the identical instruction prompt three times and determined each sentence’s relevancy label by majority vote. For the open-source models, we applied ContextCite to quantify the relevance of each sentence within the QA contexts, as ContextCite provides a simple, scalable mechanism for tracing portions of a generated response back to specific input sentences \cite{cohen-wang_contextcite_2024}. Each model received the identical prompt used by the physician labelers to elicit sentence-level relevance judgments. The complete prompts for generating self-reported labels and ContextCite annotations are provided in Appendix section \ref{LLM_Self_Report_Prompt}.

\subsection{Problem Formulation}

Suppose that a particular question consists of a set of sentences $\mathcal{S} = \mathcal{S}^+ \cup \mathcal{S}^-$, where $\mathcal{S}^+$ is the set of relevant sentences (as labeled by a physician trainee), and $\mathcal{S}^-$ is the set of irrelevant sentences. Let $Y \in \{1, 2, ..., K\}$ be the true label. Suppose we have some LLM $f: 2^{\mathcal{S}} \rightarrow \{1, 2, ..., K\}$.

In order to probe whether $f$ answers the question using the same information as a human, we compare $f(\mathcal{S})$ with $f(\mathcal{S}^+)$, under the assumption that the set $f(\mathcal{S}^+)$ is sufficient for a human to answer the question correctly. This gives us the following possible scenarios:
\begin{enumerate}
 \item $f(\mathcal{S}) = f(\mathcal{S}^+) = Y$: The model is correct in both cases, suggesting that it relies on the same information as a human to solve the problem.
    \item $f(\mathcal{S}) \neq Y,  f(\mathcal{S}^+) \neq Y$: The model is incorrect in both cases, indicating that the problem is inherently difficult or $f$ has poor capabilities.
     \item $f(\mathcal{S}) = Y,  f(\mathcal{S}^+) \neq Y$: By removing irrelevant sentences, we flip a correct prediction to an incorrect one. This indicates that the model may have been relying on spurious information in $\mathcal{S}^-$ (i.e. information for which a human deems irrelevant) to make its predictions. 
      \item $f(\mathcal{S}) \neq Y, f(\mathcal{S}^+) = Y$: The model improves when irrelevant information is removed, indicating that $\mathcal{S}^+$ contains sufficient information to answer the question as expected, and the presence of $\mathcal{S}^-$ introduces noise or distractions.
\end{enumerate}

As case (3) is the most salient, we propose a metric to evaluate $f$ based on the prevalence of samples which fall into this case. Specifically, we define the \ourmetric (\textbf{S}purious \textbf{R}ate), which is computed as:
$$\ourmetric(f) = \frac{\sum_{i=1}^N \mathbf{1}[f(\mathcal{S}_i) = Y_i \land f(\mathcal{S}_i^+) \neq Y_i]}{\sum_{i=1}^N \mathbf{1}[f(\mathcal{S}_i) = Y_i]},$$
where $N$ is the total number of questions. 
A higher \ourmetric indicates greater reliance on spurious or irrelevant information, while a lower value suggests the model's predictions are more robust to the removal of distractors and better aligned with human problem solving.

% \section{Dataset Analysis}
\subsection{Evaluation Set-Up}

\begin{figure}[h!]
  \centering
  \includegraphics[width=0.8\textwidth]{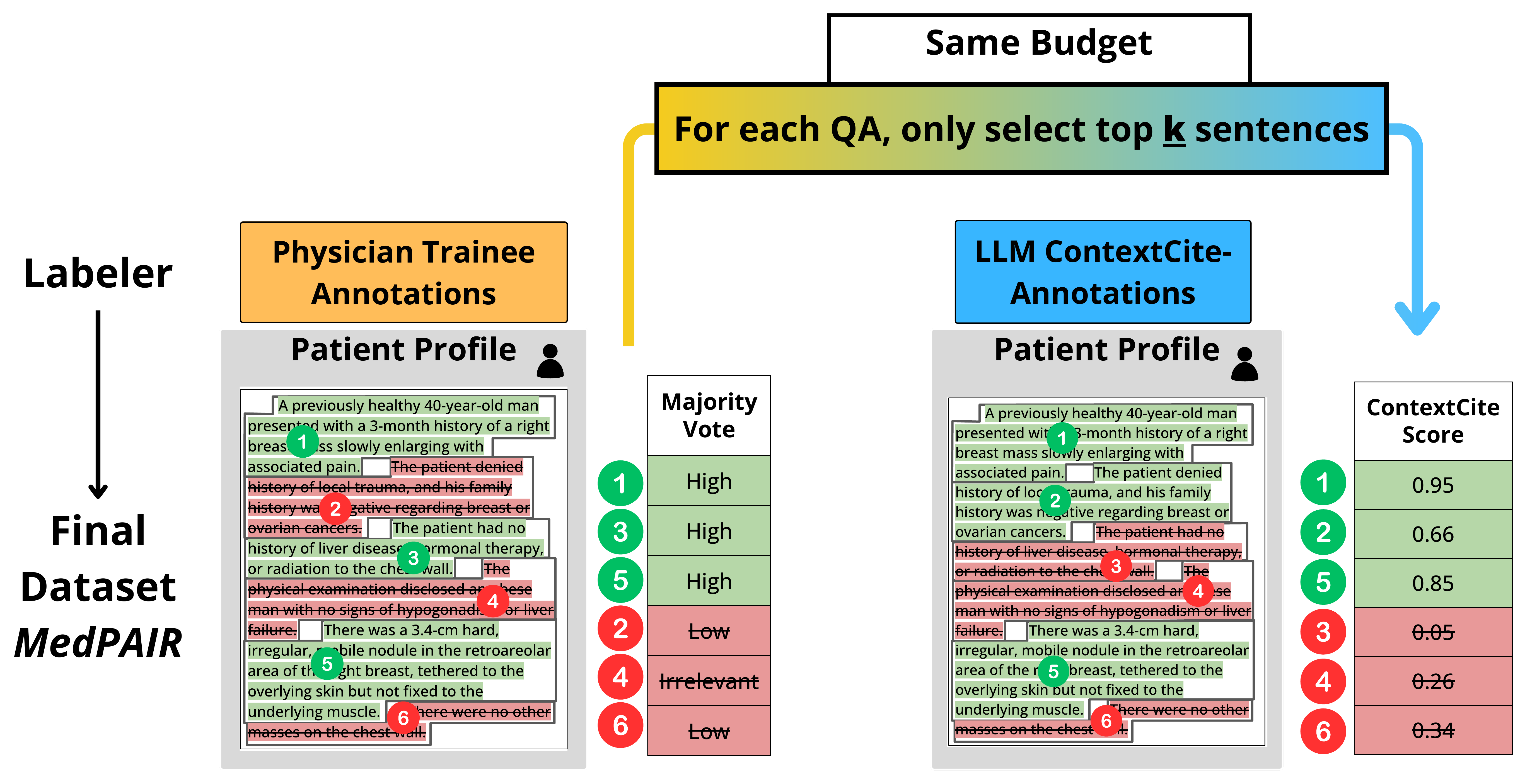}
  \caption{\textbf{Aligning Physician Trainee Annotations with LLM ContextCite Raw Scores Using an Identical Input Context Budget.}}
  \label{evaluation_Benchmark}
\end{figure}

To compare the LLM-generated ContextCite scores (numerical) with the relevance labels assigned by the physician trainee (three categories) for each sentence, we established a matching metrics between ternary labels and ContextCite scores to map the numerical scores to the categorical labels. For each QA pair, we let k equal the number of sentences marked relevant by majority vote. We then selected the k sentences with the highest raw ContextCite scores and labeled them “\textit{high relevance}.” The remaining sentences were ranked and assigned to “\textit{low relevance}” or “\textit{irrelevant}” based on their score order. This alignment creates a direct mapping between LLM attributions and human judgments, allowing us to assess how well the model’s sentence rankings match expert annotations. Figure \ref{evaluation_Benchmark} illustrates this matching process, showing how trainee‐provided labels are applied to ContextCite outputs.

\section{Results}

\subsection{Dataset Characteristics}

We received a total of 6,224 QA labels from 36 labelers and only 2,918 labels answer the step 1 classification correctly (Figure \ref{fig:study_design}). There are 1,404 unique QAs with all correct physician trainees label,
with 104 QAs which contain only highly relevant sentences and for which the QA would therefore be the same after low or irrelevant sentences removal. In the end, we curated 1,300 QAs which contained at least one removed low-relevance or irrelevant sentence. 

The four QAs have different characteristics, as JAMA Clinical Challenge are usually long and have lots of details, with each sentence containing more words on average. The low relevance and irrelevant sentences also show the characteristics in perplexity that they are harder to predict as they are more complex, less structured, or diverge from typical language patterns the model has seen during training.

% \begin{table}[]
% \begin{tabular}{ll}
%  & Accuracy Rate \\
% MMLU & 84.2 (0.5) \\
% JAMA Challenge & 55.2 (0.5) \\
% MedBullets & 65.8 (0.4) \\
% MedXpertQA & 23.5 (0.4) \\
% Overall & 48.3
% \end{tabular}
% \end{table}

We compared the final 1300 QAs \emph{MedPair} on their average of sentences, words per sentence, and perplexity. Across the four medical QA datasets, the highly relevant sentences are consistently longer and more uniform in structure, with lower perplexity values and thus greater linguistic predictability. In contrast, irrelevant or low‐relevance sentences were shorter on average, displayed much higher variability in length, and proved more difficult for the language model to anticipate. %There is a clear distinguish between the high and low/irrelevant sentences.

\begin{table}[h!]
\centering
\small
\begin{tabular}{l|c|c|c|c|c|c|c}
\toprule
\textbf{Dataset} & \textbf{Total QA} & \textbf{\makecell[c]{Total\\ Options}} & \textbf{\makecell[c]{Avg\\ Sentence}} & \multicolumn{2}{c}{\textbf{\makecell[c]{Avg Words \\Per Sentence}}} & \multicolumn{2}{c}{\textbf{Perplexity}} \\\hline
&  & & & \textbf{High} & \textbf{Low/Irr} & \textbf{High} & \textbf{Low/Irr} \\
\toprule
\midrule
\textbf{\makecell[l]{MMLU \\{\scriptsize(Precision Medicine)}}}     & 193 & 4 & 15.9 {\scriptsize(7.0)} & 18.7 {\scriptsize(5.2)}  & 12.8 {\scriptsize(4.6)} & 46.4 {\scriptsize(56.3)} & 58.7  {\scriptsize(70.4)} \\\hline
\textbf{\makecell[l]{JAMA \\ Clinical Challenge}}    & 582 & 4 & 26.8 {\scriptsize(8.5)} & 23.1 {\scriptsize(5.6)} & 16.0 {\scriptsize(5.4)} & 51.6 {\scriptsize(69.3)} & 68.2 {\scriptsize(92.4)} \\\hline
\textbf{MedBullets}    & 207 & 4 & 21.0 {\scriptsize(4.6)} & 18.1 {\scriptsize(4.2)} & 16.0 {\scriptsize(4.3)} & 46.5 {\scriptsize(51.1)} & 48.3 {\scriptsize(65.8)} \\\hline
\textbf{MedXpertQA}     & 318 & 10  & 14.9 {\scriptsize(5.6)} & 21.4 {\scriptsize(6.8)} & 15.6 {\scriptsize(4.9)} & 41.4 {\scriptsize(43.8)} & 52.3 {\scriptsize(71.0)} \\
\bottomrule
\textbf{Overall}     & 1300 & 4/10 & 21.3 {\scriptsize(8.8)} & 21.2 {\scriptsize(6.0)} & 15.4 {\scriptsize(5.1)} & 48.7 {\scriptsize(62.0)} & 61.0 {\scriptsize(82.9)} \\
\bottomrule
\end{tabular}
\vspace{0.7mm}
\caption{\textbf{Comparative Analysis of Physician Trainee–Annotated \emph{MedPair} Dataset Characteristics.} Values in parentheses represent standard deviations.}
\label{tab:comparison_1}
\end{table}

\subsection{Humans and LLMs Disagree on Information Relevance}

\begin{wraptable}{l}{0.65\textwidth}% reduce width from 0.7 to 0.6
  \centering
  \small                         % use smaller font
  \setlength{\tabcolsep}{3pt}         % tighten inter-column padding
  \begin{tabular}{@{}l|c|c|c|c@{}}     % remove extra side padding
    \toprule
    Data Source & Qwen-72B & Llama-70B & Qwen-14B & GPT-4o\\
    \midrule
                & CC & CC & CC & SR\\
    \midrule
    MMLU       & 26.9 (0.2)  & \textbf{70.7} (0.2) & 56.9 (0.2) & 50.5 (0.3)\\
    JAMA       & 45.5 (0.2)  & \textbf{62.1} (0.2) & 59.1 (0.2) & 45.2 (0.3)\\
    MedBullets & 49.8 (0.3)  & \textbf{66.6} (0.2) & 53.9 (0.2) & 45.2 (0.3)\\
    MedXpertQA & 51.8 (0.3)  & \textbf{69.3} (0.3) & 51.9 (0.2) & 52.1 (0.4)\\ 
    \midrule
    Overall    & 44.9 (0.3)  & \textbf{65.9} (0.2) & 56.2 (0.2) & 47.7 (0.3)\\
    \bottomrule
  \end{tabular}
  \caption{\textbf{Relevance Label Concordance (\%)} with Physician Trainee Labels. “CC” denotes ContextCite score; “SR” denotes Self-Reported labels. Standard deviations in parentheses.}
  \label{concordance}
\end{wraptable}

By examining cases in which physician trainees and LLMs produced differing relevance annotations, \emph{MedPAIR} reveals fundamental differences in how each identifies and priorities clinically relevant input context. We quantified the agreement between sentences marked as highly relevant by physician trainees and those highlighted by the models, using ContextCite scores for Qwen-14B, Llama-70B and Qwen-72B alongside GPT-4o self reporting. Although Llama-70B achieved the highest agreement rate at 65.9 percent, the concordance did not exceed two thirds of all instances. More than thirty percent of sentences identified as \textit{"highly relevant"} by clinicians were not recognized by the models as highly relevant. Such discrepancies in relevance annotation are likely to affect the QA accuracy. The results are shown in table \ref{concordance}.

A common pattern was overattention to superficial cues. For example, a model might latch onto a laboratory value that is extreme and assume it must be important, even if it is not relevant to the question at hand. %In one case, an LLM fixated on an incidental finding (“mild elevation in liver enzymes”) when diagnosing a skin rash, marking that sentence as relevant and incorporating it into its answer explanation – a form of distractor usage. The human annotators, by contrast, recognized that the rash diagnosis had nothing to do with the liver enzyme quirk and flagged that sentence as irrelevant. Such cases reveal how an LLM’s internal heuristics (e.g. “unusual values are important”) can mislead its reasoning, whereas human experts apply informed judgment to filter out red herrings. 
Conversely, models sometimes missed subtle but crucial cues that humans tagged as relevant. %For instance, a single sentence mentioning “\textit{recent travel to the Ohio River Valley}” in a patient’s history is a critical clue for a fungal infection diagnosis, physician labelers spot it immediately as a key piece of context, but a LLM might ignore it if it hasn’t connected that region with certain infections in its knowledge. These misses show that even highly knowledgeable LLMs may fail to connect the dots in context, especially when the clue requires integrating world knowledge in a non-obvious way.
These findings could be partially due to LLMs occasionally attribute incorrect answers on misinterpreted or irrelevant context, indicating flawed input context relevance estimates. ContextCite highlights cases where a model justifies its answer by citing a sentence it wrongly deems supportive, what researchers term contributive attribution. %For instance, a model might recommend an inappropriate treatment due to a misread clinical detail. These discrepancies between model-perceived and expert-validated relevance expose critical gaps in reasoning.

\subsection{Human Relevance Improves LLM Performance}
% We find that MORE HERE ON HOW MODELS ARE BETTER WHEN THEY USE HUMAN RELEVANCE BUT NOT LLM OR CONTEXTCITE? 

After removing low-relevance and irrelevant sentences, LLM performance improved when limited to the physician trainee majority-vote labeled sentences marked as highly relevant. This filtering effectively constrained the model’s attention to clinically pertinent information, reducing the noise introduced by less relevant context. Physician labeling instructions explicitly emphasized that QA tasks could be completed using only these highly relevant sentences, ensuring that models concentrated on the critical details necessary for accurate decision-making. 

\begin{figure}[h!]
  \centering
  \includegraphics[width=\textwidth]{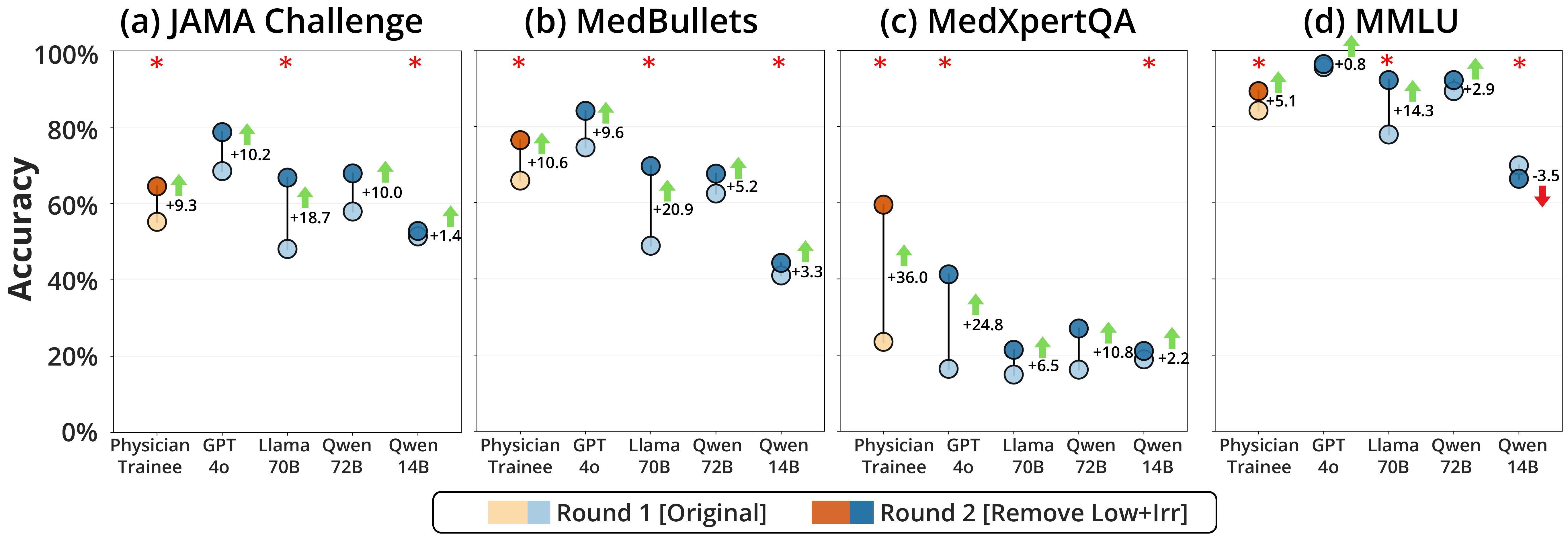}
  \small * denotes that statistical significance p < 0.001.
  \caption{\textbf{Effect of Filtering Context on Final Performance.}  GPT-4o outperforms all tested open-source language models. After removing irrelevant and low-relevance sentences, LLaMA 70B and Qwen 14B demonstrated the most substantial accuracy improvements. In contrast, Qwen 72B occasionally experiences performance drops following the removal process. }
  \label{results}
\end{figure}

\begin{table}[b]
  \centering
  \small
  \begin{tabular}{l|c|c|c|c}
    \toprule
    \textbf{Models} & \textbf{MMLU} & \textbf{JAMA} 
      & \textbf{MedBullets} & \textbf{MedXpertQA} \\
    \midrule
    Llama-70B   & 1.6  & 8.9  & 7.7  & 6.0 \\
    Qwen-72B    & 2.6  & 8.6  & 5.8  & 4.1 \\
    Qwen-14B    & 9.8  & 13.9 & 18.5 & 8.8 \\
    GPT-4o      & 2.1  & 6.5  & 4.8  & 4.1 \\
    \bottomrule
  \end{tabular}
  \vspace{1mm}
  \caption{\textbf{The \ourmetric(\%) of removing physician trainee-identified low-relevance and irrelevant sentences.} Each number denotes the proportion of questions that were answered correctly in Round 1 but became incorrect in Round 2 after those sentences were removed.}
  \label{tab:spurious_rate}
\end{table}

Figure \ref{results} demonstrates that excising sentences deemed low-relevance by physician trainees yields substantial accuracy gains for most LLMs. Noted that in round 2, physician trainee only annotated 248 QAs (the same sampling ratio for each dataset as 1,300 QAs). Notably, Qwen-72B’s accuracy increases from 89.3\% to 92.2\% on the MMLU Precision Medicine subset and from 35.1\% to 62.2\% overall, while GPT-4o improves from 95.6\% to 96.4\% and from 39.3\% to 73.0\%, respectively, preserving its position as the highest-performing model before and after filtering. Parallel improvements appear on the JAMA Clinical Challenge, MedBullets, and MedXpertQA datasets, with standard deviations remaining under 0.5 in nearly every case, indicating consistent benefits of relevance pruning. In contrast, Qwen-14B and Llama-70B exhibit modest declines on the MMLU subset—marked in red—suggesting that less advanced models may sometimes rely on information classified as irrelevant. Overall, these findings underscore that expert-guided sentence removal can markedly enhance LLM performance in clinical QA, even surpassing the unfiltered accuracy of physician trainees (48.3\%).

\pgfplotstableset{
  heatmap/.style={
    color cells={
      min=-16.9,
      max=24.8,
      textcolor=black,
      colormap/blues
    },
    every head row/.style={before row=\toprule, after row=\midrule},
    every last row/.style={after row=\bottomrule}
  }
}

While the performance gains were modest, the results indicate that focusing on high-relevance input enables the models to avoid distractions from extraneous information that could otherwise skew their predictions. This targeted approach demonstrates the value of fine-grained relevance curation in enhancing LLM decision-making reliability in clinical contexts. As shown in Table \ref{tab:spurious_rate}, there are a subset of QAs (ranging from 1.6\% - 18.5\%) which LLM depend on sentences annotated as low-relevance or irrelevant to arrive at the correct answer. Among the models evaluated, Qwen-14B exhibits the highest \ourmetric, while the closed-source GPT-4o exhibits the lowest.

\subsection{LLM Relevance Improves LLM Performance}

The disagreement between physician trainees and LLMs on the input context relevancy reveals the differences in highly relevant sentences. To assess how these differences influence model accuracy, we pruned question contexts according to four criteria: the original unaltered text; sentences retained by physician trainees; sentences retained by Qwen-72B and Llama-70B via ContextCite scoring; and sentences self-reported as relevant by GPT-4o. We then re‐evaluated GPT-4o on each reduced context, as it performs the best on the original data. 

% optional helper macro
\newcommand{\rheat}[2]{\cellcolor{red!#1}{#2}}
\newcommand{\heat}[2]{\cellcolor{cyan!#1}{#2}}

\begin{table}[htbp]
  \centering
  \small
  \begin{tabular}{lcccc}
    \toprule
    \textbf{Datasets} 
      & \textbf{MMLU} & \textbf{JAMA} 
      & \textbf{MedBullets} & \textbf{MedXpertQA} \\
    \midrule
    Original
      & 95.6   & 68.5   & 74.5   & 16.4   \\
    \midrule
    \makecell[l]{After \textbf{Physician Trainee}\\Labeled Low+Irr Removal}
      & \rheat{5}{+0.8}    & \rheat{30}{+10.2}
      & \rheat{35}{+9.6}   & \rheat{65}{+24.8}  \\
    \midrule
    \makecell[l]{After \textbf{Qwen-72B}\\Low+Irr Removal}
      & \heat{10}{\(-\)1.8} & \rheat{15}{+4.0}
      & \rheat{15}{+2.3}   & \rheat{65}{+24.6}  \\
    \midrule
    \makecell[l]{After \textbf{Llama-70B}\\Low+Irr Removal}
      & \heat{30}{\(-\)2.4} & \rheat{5}{+0.7}
      & \rheat{5}{+0.1}    & \rheat{55}{+22.4}  \\
    \midrule
    \makecell[l]{After \textbf{GPT-4o} Self-Reported\\Low+Irr Removal}
      & \rheat{10}{+1.8}   & \rheat{35}{+10.4}
      & \rheat{20}{+8.6}   & \rheat{20}{+8.8}   \\
    \bottomrule
  \end{tabular}
  \vspace{0.2mm}
  \caption{\textbf{Heatmap of GPT-4o performance gains (\%).} Red shades denote positive gains; blue shades denote losses.}
  \label{tab:heatmap_standard}
\end{table}

  \vspace{-2.2mm}

In smaller benchmarks such as MMLU, pruning based on non–GPT-4o criteria sometimes led to modest accuracy declines. By contrast, every pruning strategy yielded dramatic gains on MedXpertQA—where shorter average contexts and a larger answer set amplify the benefit of removing irrelevant material—boosting accuracy by 22.4\% to 24.8\%. The largest improvement occurred with physician-curated pruning, while ContextCite-based selection from Qwen-72B and Llama-70B delivered moderate gains. GPT-4o’s own self-reported labels proved the least reliable, occasionally degrading performance. These findings underscore the superior value of expert human judgments for relevance curation in clinical question answering.

\subsection{Qualitative Results}

A board-certified physician, HJ, reviewed the physician-annotated majority-vote outcomes. Analysis of high- and low-relevance labels reveals that text marked as highly relevant by the physician trainee contains more anatomical structures and comparative descriptions (e.g., progressive, increased), whereas low-relevance text includes more historical information (medication, allergy, travel, social (smoking, illicit drug), etc.) and negative findings (uncomplicated, noncontributory, etc.)(Table \ref{tab:keyword_readability}).

From the full dataset, 30 QA pairs were randomly selected and Dr. HJ compared original and edited versions after removing irrelevant sentences, then categorized these removed low relevance or irrelevant sentences into thematic groups such as \textit{1) Redundant Clinical Details}, \textit{2) Negative Result} that is not essential for current chief complaint, \textit{3) Low relevant or Irrelevant Temporal Information}, \textit{4) History (Medical, Surgical, Medication, Social) with No/Very Low Clinical Information}, etc. The validation exercise evaluated whether the remaining highly relevant sentences maintained the link to the correct answer and whether removing low-relevance content affected answer correctness. The sample case study is presented in Appendix section \ref{sample} and the validation sheet is available in the supplementary material.

\section{Discussion}
Our findings highlight a significant mismatch between LLM and human expert estimated relevance in the evaluation of clinical vignettes. This resonates with concerns raised in earlier work \cite{bao_llms_2024} that LLM performance can be overestimated if one only looks at accuracy \cite{ishii_analysis_2024}. Such discordance suggests that accuracy metrics alone may fail to capture how large language models derive answers from clinical context. %Matching LLM's relevance estimates with domain experts could enhance interpretability and support more actionable output.
Alignment between model-assigned and physician-assigned relevance is essential for developing clinically deployable AI, where safe and effective integration depends not only on producing accurate outputs but also on correct interpretation on the input context. Models that prioritize the same clinically meaningful information as human experts are more likely to support interpretable and actionable decision-making. 
Previous work has demonstrated that selectively pruning input contexts and retaining only the most relevant context can enhance QA performance in language models \cite{mckechnie_context_2025, kabongo_effective_2024}. Our experiments extend these findings by showing that context reduction guided by physician annotations, ContextCite scores from open‐source models, or few‐shot self‐report prompting of GPT-4o each provides consistent performance gains across four medical QAs. %\emph{MedPAIR} addresses this need by introducing a fine-grained, expert-annotated benchmark for comparing model and physician relevance assessments. By establishing an explicit physician-LLM relevance pair at the sentence level for each clinical case, \emph{MedPAIR} enables direct comparison of reasoning processes and offers a foundation for evaluating how LLMs interpret and prioritize clinical information, an essential step toward developing models suitable for clinical deployment. 

% Our study asks a fundamental question to instead of looking at the results discrepancies, we choose to look at the initial information before LLM starts to make predictions or reasonings. 

%LLM as a judge
Additionally, the \emph{MedPAIR} dataset contributes to understand whether LLM is able to automate the evaluation process as a judge. While our findings suggest that LLMs can enhance performance in this role, the substantial improvements observed with domain expert-generated datasets demonstrate the importance of human involvement in the evaluation process \cite{shi_judging_2025, zheng_judging_2023}. The human and LLM disagreements on information relevance highlight the need for expert oversight in ensuring accuracy \cite{soboroff_dont_2025}. Although LLMs can provide ContextCite scores and self-report labels to explain the identification of input context, the quality and consistency of these outputs still require validation from human experts. This is particularly important in healthcare, where automating prediction and evaluation with LLMs could have serious consequences due to potential misalignments with human judgment in input information retrieval \cite{abbasiantaeb_can_2024}.

\section{Limitation \& Future Work}

% While our intervention improved alignment, it is a simplified setup. We had the luxury of human annotations for each question, which is not available in real time for novel cases. In practice, one might deploy a two-step system: first have the model or another auxiliary model predict what parts of a case are relevant (perhaps via a model trained on our annotated data), then feed those to the main model. This could approximate our guided approach without needing a human in the loop every time. 

Interpreting LLM's input relevance scoring using ContextCite scores and self-reported labels may lack reliability \cite{gu_survey_2025, panickssery_llm_2024}. ContextCite scores do not always accurately capture the relevance of each sentence in decision-making for question answering, while self-reported labels are often inconsistent and may not align with actual annotations. It's critical to understand how LLMs evaluate sentence relevance within patient profiles and new evaluation metrics or measurement approaches may be necessary. Given that human interpretations are costly and time-consuming, we are limited to a small subset of data, which restricts the ability to ensure generalizability within a larger alignment framework. In addition, while removing irrelevant and low-relevance sentences improved accuracy, relying solely on human annotations for this task is impractical for real-time clinical scenarios \cite{croxford_current_2025, sharma_relevance_2017}. Moving forward, we aim to use physician-in-the-loop \emph{MedPAIR} benchmark to fine-tune text-based LLMs (e.g., Llama-3 and Mistral), aligning their contextual relevance judgments more closely with physician reasoning. This enhanced alignment is expected to significantly improve LLM performance in medical QA tasks by enabling models to prioritize clinically relevant information effectively.

\section{Conclusion}

The \emph{MedPAIR} benchmark establishes a rigorous pre-reasoning evaluation by quantifying sentence-level alignment between LLM relevance judgments and physician‐trainee annotations across a comprehensive suite of medical QA scenarios. We introduce the notion of relevance pairs, highlighting which parts of a problem should be central to solving it, and used these maps to diagnose mismatches in how an AI approaches clinical reasoning. Our experiments with 1,300 annotated QA examples revealed that, although the LLM can arrive at correct answers, by solely focusing on the physician-labeled highly relevant input context, LLM performance can be improved. The \emph{MedPAIR} benchmark lays the groundwork for developing LLMs whose performance meet the exacting demands of real‐world medical practice. %We anticipate this benchmark will support the development of more generalizable, interpretable, and trustworthy reasoning across diverse biomedical applications.

\section{Acknowledgement}

This work was supported in part by an award from the Hasso Plattner Foundation, a National Science Foundation (NSF) CAREER Award (\#2339381), and an AI2050 Early Career Fellowship (G-25-68042).

\bibliographystyle{plain}
\bibliography{references}

\begin{thebibliography}{10}

\bibitem{abbasiantaeb_can_2024}
Zahra Abbasiantaeb, Chuan Meng, Leif Azzopardi, and Mohammad Aliannejadi.
\newblock Can {We} {Use} {Large} {Language} {Models} to {Fill} {Relevance} {Judgment} {Holes}?, May 2024.
\newblock arXiv:2405.05600 [cs].

\bibitem{adlakha_evaluating_2024}
Vaibhav Adlakha, Parishad BehnamGhader, Xing~Han Lu, Nicholas Meade, and Siva Reddy.
\newblock Evaluating {Correctness} and {Faithfulness} of {Instruction}-{Following} {Models} for {Question} {Answering}.
\newblock {\em Transactions of the Association for Computational Linguistics}, 12:681--699, 2024.
\newblock Place: Cambridge, MA Publisher: MIT Press.

\bibitem{ahn_prompt-reverse_2025}
Jihyun~Janice Ahn and Wenpeng Yin.
\newblock Prompt-{Reverse} {Inconsistency}: {LLM} {Self}-{Inconsistency} {Beyond} {Generative} {Randomness} and {Prompt} {Paraphrasing}, April 2025.
\newblock arXiv:2504.01282 [cs] version: 1.

\bibitem{atil_llm_2024}
Berk Atil, Alexa Chittams, Liseng Fu, Ferhan Ture, Lixinyu Xu, and Breck Baldwin.
\newblock {LLM} {Stability}: {A} detailed analysis with some surprises.
\newblock {\em CoRR}, January 2024.

\bibitem{bansal_does_2021}
Gagan Bansal, Tongshuang Wu, Joyce Zhou, Raymond Fok, Besmira Nushi, Ece Kamar, Marco~Tulio Ribeiro, and Daniel Weld.
\newblock Does the {Whole} {Exceed} its {Parts}? {The} {Effect} of {AI} {Explanations} on {Complementary} {Team} {Performance}.
\newblock In {\em Proceedings of the 2021 {CHI} {Conference} on {Human} {Factors} in {Computing} {Systems}}, {CHI} '21, pages 1--16, New York, NY, USA, May 2021. Association for Computing Machinery.

\bibitem{bao_llms_2024}
Guangsheng Bao, Hongbo Zhang, Linyi Yang, Cunxiang Wang, and Yue Zhang.
\newblock {LLMs} with {Chain}-of-{Thought} {Are} {Non}-{Causal} {Reasoners}.
\newblock {\em CoRR}, January 2024.

\bibitem{bohnet_attributed_2023}
Bernd Bohnet, Vinh~Q. Tran, Pat Verga, Roee Aharoni, Daniel Andor, Livio~Baldini Soares, Massimiliano Ciaramita, Jacob Eisenstein, Kuzman Ganchev, Jonathan Herzig, Kai Hui, Tom Kwiatkowski, Ji~Ma, Jianmo Ni, Lierni~Sestorain Saralegui, Tal Schuster, William~W. Cohen, Michael Collins, Dipanjan Das, Donald Metzler, Slav Petrov, and Kellie Webster.
\newblock Attributed {Question} {Answering}: {Evaluation} and {Modeling} for {Attributed} {Large} {Language} {Models}, February 2023.
\newblock arXiv:2212.08037 [cs].

\bibitem{bucinca_trust_2021}
Zana Buçinca, Maja~Barbara Malaya, and Krzysztof~Z. Gajos.
\newblock To {Trust} or to {Think}: {Cognitive} {Forcing} {Functions} {Can} {Reduce} {Overreliance} on {AI} in {AI}-assisted {Decision}-making.
\newblock {\em Proc. ACM Hum.-Comput. Interact.}, 5(CSCW1):188:1--188:21, April 2021.

\bibitem{cabral_clinical_2024}
Stephanie Cabral, Daniel Restrepo, Zahir Kanjee, Philip Wilson, Byron Crowe, Raja-Elie Abdulnour, and Adam Rodman.
\newblock Clinical {Reasoning} of a {Generative} {Artificial} {Intelligence} {Model} {Compared} {With} {Physicians}.
\newblock {\em JAMA Internal Medicine}, 184(5):581--583, May 2024.

\bibitem{chanda_dermatologist-like_2024}
Tirtha Chanda, Katja Hauser, Sarah Hobelsberger, Tabea-Clara Bucher, Carina~Nogueira Garcia, Christoph Wies, Harald Kittler, Philipp Tschandl, Cristian Navarrete-Dechent, Sebastian Podlipnik, Emmanouil Chousakos, Iva Crnaric, Jovana Majstorovic, Linda Alhajwan, Tanya Foreman, Sandra Peternel, Sergei Sarap, İrem Özdemir, Raymond~L. Barnhill, Mar Llamas-Velasco, Gabriela Poch, Sören Korsing, Wiebke Sondermann, Frank~Friedrich Gellrich, Markus~V. Heppt, Michael Erdmann, Sebastian Haferkamp, Konstantin Drexler, Matthias Goebeler, Bastian Schilling, Jochen~S. Utikal, Kamran Ghoreschi, Stefan Fröhling, Eva Krieghoff-Henning, and Titus~J. Brinker.
\newblock Dermatologist-like explainable {AI} enhances trust and confidence in diagnosing melanoma.
\newblock {\em Nature Communications}, 15(1):524, January 2024.
\newblock Publisher: Nature Publishing Group.

\bibitem{chen_benchmarking_2025}
Hanjie Chen, Zhouxiang Fang, Yash Singla, and Mark Dredze.
\newblock Benchmarking {Large} {Language} {Models} on {Answering} and {Explaining} {Challenging} {Medical} {Questions}.
\newblock In Luis Chiruzzo, Alan Ritter, and Lu~Wang, editors, {\em Proceedings of the 2025 {Conference} of the {Nations} of the {Americas} {Chapter} of the {Association} for {Computational} {Linguistics}: {Human} {Language} {Technologies} ({Volume} 1: {Long} {Papers})}, pages 3563--3599, Albuquerque, New Mexico, April 2025. Association for Computational Linguistics.

\bibitem{chen_reasoning_nodate}
Yanda Chen, Joe Benton, Ansh Radhakrishnan, Jonathan Uesato~Carson Denison, John Schulman, Arushi Somani, Peter Hase, Misha Wagner Fabien Roger~Vlad Mikulik, Sam Bowman, Jan Leike~Jared Kaplan, and {others}.
\newblock Reasoning {Models} {Don}’t {Always} {Say} {What} {They} {Think}.

\bibitem{chiang_can_2023}
Cheng-Han Chiang and Hung-yi Lee.
\newblock Can {Large} {Language} {Models} {Be} an {Alternative} to {Human} {Evaluations}?
\newblock In Anna Rogers, Jordan Boyd-Graber, and Naoaki Okazaki, editors, {\em Proceedings of the 61st {Annual} {Meeting} of the {Association} for {Computational} {Linguistics} ({Volume} 1: {Long} {Papers})}, pages 15607--15631, Toronto, Canada, July 2023. Association for Computational Linguistics.

\bibitem{choe_what_2024}
Sang~Keun Choe, Hwijeen Ahn, Juhan Bae, Kewen Zhao, Minsoo Kang, Youngseog Chung, Adithya Pratapa, Willie Neiswanger, Emma Strubell, Teruko Mitamura, Jeff Schneider, Eduard Hovy, Roger Grosse, and Eric Xing.
\newblock What is {Your} {Data} {Worth} to {GPT}? {LLM}-{Scale} {Data} {Valuation} with {Influence} {Functions}, May 2024.
\newblock arXiv:2405.13954 [cs].

\bibitem{choi_identifying_2024}
Jaekeol Choi.
\newblock Identifying {Key} {Terms} in {Prompts} for {Relevance} {Evaluation} with {GPT} {Models}, May 2024.
\newblock arXiv:2405.06931 [cs].

\bibitem{chuang_selfcite_2025}
Yung-Sung Chuang, Benjamin Cohen-Wang, Shannon~Zejiang Shen, Zhaofeng Wu, Hu~Xu, Xi~Victoria Lin, James Glass, Shang-Wen Li, and Wen-tau Yih.
\newblock {SelfCite}: {Self}-{Supervised} {Alignment} for {Context} {Attribution} in {Large} {Language} {Models}, February 2025.
\newblock arXiv:2502.09604 [cs].

\bibitem{cohen-wang_learning_2025}
Benjamin Cohen-Wang, Yung-Sung Chuang, and Aleksander Madry.
\newblock Learning to {Attribute} with {Attention}, April 2025.
\newblock arXiv:2504.13752 [cs].

\bibitem{cohen-wang_contextcite_2024}
Benjamin Cohen-Wang, Harshay Shah, Kristian Georgiev, and Aleksander Madry.
\newblock {ContextCite}: {Attributing} {Model} {Generation} to {Context}.
\newblock November 2024.

\bibitem{croxford_current_2025}
Emma Croxford, Yanjun Gao, Nicholas Pellegrino, Karen Wong, Graham Wills, Elliot First, Frank Liao, Cherodeep Goswami, Brian Patterson, and Majid Afshar.
\newblock Current and future state of evaluation of large language models for medical summarization tasks.
\newblock {\em npj Health Systems}, 2(1):1--13, February 2025.
\newblock Publisher: Nature Publishing Group.

\bibitem{cutler_physician_2019}
David Cutler, Jonathan~S. Skinner, Ariel~Dora Stern, and David Wennberg.
\newblock Physician {Beliefs} and {Patient} {Preferences}: {A} {New} {Look} at {Regional} {Variation} in {Health} {Care} {Spending}.
\newblock {\em American Economic Journal: Economic Policy}, 11(1):192--221, February 2019.

\bibitem{deyoung_eraser_2020}
Jay DeYoung, Sarthak Jain, Nazneen~Fatema Rajani, Eric Lehman, Caiming Xiong, Richard Socher, and Byron~C. Wallace.
\newblock {ERASER}: {A} {Benchmark} to {Evaluate} {Rationalized} {NLP} {Models}.
\newblock In Dan Jurafsky, Joyce Chai, Natalie Schluter, and Joel Tetreault, editors, {\em Proceedings of the 58th {Annual} {Meeting} of the {Association} for {Computational} {Linguistics}}, pages 4443--4458, Online, July 2020. Association for Computational Linguistics.

\bibitem{es_ragas_2024}
Shahul Es, Jithin James, Luis Espinosa~Anke, and Steven Schockaert.
\newblock {RAGAs}: {Automated} {Evaluation} of {Retrieval} {Augmented} {Generation}.
\newblock In Nikolaos Aletras and Orphee De~Clercq, editors, {\em Proceedings of the 18th {Conference} of the {European} {Chapter} of the {Association} for {Computational} {Linguistics}: {System} {Demonstrations}}, pages 150--158, St. Julians, Malta, March 2024. Association for Computational Linguistics.

\bibitem{farquhar_detecting_2024}
Sebastian Farquhar, Jannik Kossen, Lorenz Kuhn, and Yarin Gal.
\newblock Detecting hallucinations in large language models using semantic entropy.
\newblock {\em Nature}, 630(8017):625--630, June 2024.
\newblock Publisher: Nature Publishing Group.

\bibitem{funkquist_citebench_2023}
Martin Funkquist, Ilia Kuznetsov, Yufang Hou, and Iryna Gurevych.
\newblock {CiteBench}: {A} {Benchmark} for {Scientific} {Citation} {Text} {Generation}.
\newblock In Houda Bouamor, Juan Pino, and Kalika Bali, editors, {\em Proceedings of the 2023 {Conference} on {Empirical} {Methods} in {Natural} {Language} {Processing}}, pages 7337--7353, Singapore, December 2023. Association for Computational Linguistics.

\bibitem{gao_enabling_2023}
Tianyu Gao, Howard Yen, Jiatong Yu, and Danqi Chen.
\newblock Enabling {Large} {Language} {Models} to {Generate} {Text} with {Citations}.
\newblock In Houda Bouamor, Juan Pino, and Kalika Bali, editors, {\em Proceedings of the 2023 {Conference} on {Empirical} {Methods} in {Natural} {Language} {Processing}}, pages 6465--6488, Singapore, December 2023. Association for Computational Linguistics.

\bibitem{gaube_non-task_2023}
Susanne Gaube, Harini Suresh, Martina Raue, Eva Lermer, Timo~K. Koch, Matthias F.~C. Hudecek, Alun~D. Ackery, Samir~C. Grover, Joseph~F. Coughlin, Dieter Frey, Felipe~C. Kitamura, Marzyeh Ghassemi, and Errol Colak.
\newblock Non-task expert physicians benefit from correct explainable {AI} advice when reviewing {X}-rays.
\newblock {\em Scientific Reports}, 13(1):1383, January 2023.
\newblock Publisher: Nature Publishing Group.

\bibitem{grattafiori_llama_2024}
Aaron Grattafiori, Abhimanyu Dubey, Abhinav Jauhri, Abhinav Pandey, Abhishek Kadian, Ahmad Al-Dahle, Aiesha Letman, Akhil Mathur, Alan Schelten, Alex Vaughan, Amy Yang, Angela Fan, Anirudh Goyal, Anthony Hartshorn, Aobo Yang, Archi Mitra, Archie Sravankumar, Artem Korenev, Arthur Hinsvark, Arun Rao, Aston Zhang, Aurelien Rodriguez, Austen Gregerson, Ava Spataru, Baptiste Roziere, Bethany Biron, Binh Tang, Bobbie Chern, Charlotte Caucheteux, Chaya Nayak, Chloe Bi, Chris Marra, Chris McConnell, Christian Keller, Christophe Touret, Chunyang Wu, Corinne Wong, Cristian~Canton Ferrer, Cyrus Nikolaidis, Damien Allonsius, Daniel Song, Danielle Pintz, Danny Livshits, Danny Wyatt, David Esiobu, Dhruv Choudhary, Dhruv Mahajan, Diego Garcia-Olano, Diego Perino, Dieuwke Hupkes, Egor Lakomkin, Ehab AlBadawy, Elina Lobanova, Emily Dinan, Eric~Michael Smith, Filip Radenovic, Francisco Guzmán, Frank Zhang, Gabriel Synnaeve, Gabrielle Lee, Georgia~Lewis Anderson, Govind Thattai, Graeme Nail, Gregoire Mialon, Guan Pang,
  Guillem Cucurell, Hailey Nguyen, Hannah Korevaar, Hu~Xu, Hugo Touvron, Iliyan Zarov, Imanol~Arrieta Ibarra, Isabel Kloumann, Ishan Misra, Ivan Evtimov, Jack Zhang, Jade Copet, Jaewon Lee, Jan Geffert, Jana Vranes, Jason Park, Jay Mahadeokar, Jeet Shah, Jelmer van~der Linde, Jennifer Billock, Jenny Hong, Jenya Lee, Jeremy Fu, Jianfeng Chi, Jianyu Huang, Jiawen Liu, Jie Wang, Jiecao Yu, Joanna Bitton, Joe Spisak, Jongsoo Park, Joseph Rocca, Joshua Johnstun, Joshua Saxe, Junteng Jia, Kalyan~Vasuden Alwala, Karthik Prasad, Kartikeya Upasani, Kate Plawiak, Ke~Li, Kenneth Heafield, Kevin Stone, Khalid El-Arini, Krithika Iyer, Kshitiz Malik, Kuenley Chiu, Kunal Bhalla, Kushal Lakhotia, Lauren Rantala-Yeary, Laurens van~der Maaten, Lawrence Chen, Liang Tan, Liz Jenkins, Louis Martin, Lovish Madaan, Lubo Malo, Lukas Blecher, Lukas Landzaat, Luke~de Oliveira, Madeline Muzzi, Mahesh Pasupuleti, Mannat Singh, Manohar Paluri, Marcin Kardas, Maria Tsimpoukelli, Mathew Oldham, Mathieu Rita, Maya Pavlova, Melanie Kambadur,
  Mike Lewis, Min Si, Mitesh~Kumar Singh, Mona Hassan, Naman Goyal, Narjes Torabi, Nikolay Bashlykov, Nikolay Bogoychev, Niladri Chatterji, Ning Zhang, Olivier Duchenne, Onur Çelebi, Patrick Alrassy, Pengchuan Zhang, Pengwei Li, Petar Vasic, Peter Weng, Prajjwal Bhargava, Pratik Dubal, Praveen Krishnan, Punit~Singh Koura, Puxin Xu, Qing He, Qingxiao Dong, Ragavan Srinivasan, Raj Ganapathy, Ramon Calderer, Ricardo~Silveira Cabral, Robert Stojnic, Roberta Raileanu, Rohan Maheswari, Rohit Girdhar, Rohit Patel, Romain Sauvestre, Ronnie Polidoro, Roshan Sumbaly, Ross Taylor, Ruan Silva, Rui Hou, Rui Wang, Saghar Hosseini, Sahana Chennabasappa, Sanjay Singh, Sean Bell, Seohyun~Sonia Kim, Sergey Edunov, Shaoliang Nie, Sharan Narang, Sharath Raparthy, Sheng Shen, Shengye Wan, Shruti Bhosale, Shun Zhang, Simon Vandenhende, Soumya Batra, Spencer Whitman, Sten Sootla, Stephane Collot, Suchin Gururangan, Sydney Borodinsky, Tamar Herman, Tara Fowler, Tarek Sheasha, Thomas Georgiou, Thomas Scialom, Tobias Speckbacher,
  Todor Mihaylov, Tong Xiao, Ujjwal Karn, Vedanuj Goswami, Vibhor Gupta, Vignesh Ramanathan, Viktor Kerkez, Vincent Gonguet, Virginie Do, Vish Vogeti, Vítor Albiero, Vladan Petrovic, Weiwei Chu, Wenhan Xiong, Wenyin Fu, Whitney Meers, Xavier Martinet, Xiaodong Wang, Xiaofang Wang, Xiaoqing~Ellen Tan, Xide Xia, Xinfeng Xie, Xuchao Jia, Xuewei Wang, Yaelle Goldschlag, Yashesh Gaur, Yasmine Babaei, Yi~Wen, Yiwen Song, Yuchen Zhang, Yue Li, Yuning Mao, Zacharie~Delpierre Coudert, Zheng Yan, Zhengxing Chen, Zoe Papakipos, Aaditya Singh, Aayushi Srivastava, Abha Jain, Adam Kelsey, Adam Shajnfeld, Adithya Gangidi, Adolfo Victoria, Ahuva Goldstand, Ajay Menon, Ajay Sharma, Alex Boesenberg, Alexei Baevski, Allie Feinstein, Amanda Kallet, Amit Sangani, Amos Teo, Anam Yunus, Andrei Lupu, Andres Alvarado, Andrew Caples, Andrew Gu, Andrew Ho, Andrew Poulton, Andrew Ryan, Ankit Ramchandani, Annie Dong, Annie Franco, Anuj Goyal, Aparajita Saraf, Arkabandhu Chowdhury, Ashley Gabriel, Ashwin Bharambe, Assaf Eisenman, Azadeh
  Yazdan, Beau James, Ben Maurer, Benjamin Leonhardi, Bernie Huang, Beth Loyd, Beto~De Paola, Bhargavi Paranjape, Bing Liu, Bo~Wu, Boyu Ni, Braden Hancock, Bram Wasti, Brandon Spence, Brani Stojkovic, Brian Gamido, Britt Montalvo, Carl Parker, Carly Burton, Catalina Mejia, Ce~Liu, Changhan Wang, Changkyu Kim, Chao Zhou, Chester Hu, Ching-Hsiang Chu, Chris Cai, Chris Tindal, Christoph Feichtenhofer, Cynthia Gao, Damon Civin, Dana Beaty, Daniel Kreymer, Daniel Li, David Adkins, David Xu, Davide Testuggine, Delia David, Devi Parikh, Diana Liskovich, Didem Foss, Dingkang Wang, Duc Le, Dustin Holland, Edward Dowling, Eissa Jamil, Elaine Montgomery, Eleonora Presani, Emily Hahn, Emily Wood, Eric-Tuan Le, Erik Brinkman, Esteban Arcaute, Evan Dunbar, Evan Smothers, Fei Sun, Felix Kreuk, Feng Tian, Filippos Kokkinos, Firat Ozgenel, Francesco Caggioni, Frank Kanayet, Frank Seide, Gabriela~Medina Florez, Gabriella Schwarz, Gada Badeer, Georgia Swee, Gil Halpern, Grant Herman, Grigory Sizov, Guangyi, Zhang, Guna
  Lakshminarayanan, Hakan Inan, Hamid Shojanazeri, Han Zou, Hannah Wang, Hanwen Zha, Haroun Habeeb, Harrison Rudolph, Helen Suk, Henry Aspegren, Hunter Goldman, Hongyuan Zhan, Ibrahim Damlaj, Igor Molybog, Igor Tufanov, Ilias Leontiadis, Irina-Elena Veliche, Itai Gat, Jake Weissman, James Geboski, James Kohli, Janice Lam, Japhet Asher, Jean-Baptiste Gaya, Jeff Marcus, Jeff Tang, Jennifer Chan, Jenny Zhen, Jeremy Reizenstein, Jeremy Teboul, Jessica Zhong, Jian Jin, Jingyi Yang, Joe Cummings, Jon Carvill, Jon Shepard, Jonathan McPhie, Jonathan Torres, Josh Ginsburg, Junjie Wang, Kai Wu, Kam~Hou U, Karan Saxena, Kartikay Khandelwal, Katayoun Zand, Kathy Matosich, Kaushik Veeraraghavan, Kelly Michelena, Keqian Li, Kiran Jagadeesh, Kun Huang, Kunal Chawla, Kyle Huang, Lailin Chen, Lakshya Garg, Lavender A, Leandro Silva, Lee Bell, Lei Zhang, Liangpeng Guo, Licheng Yu, Liron Moshkovich, Luca Wehrstedt, Madian Khabsa, Manav Avalani, Manish Bhatt, Martynas Mankus, Matan Hasson, Matthew Lennie, Matthias Reso, Maxim
  Groshev, Maxim Naumov, Maya Lathi, Meghan Keneally, Miao Liu, Michael~L. Seltzer, Michal Valko, Michelle Restrepo, Mihir Patel, Mik Vyatskov, Mikayel Samvelyan, Mike Clark, Mike Macey, Mike Wang, Miquel~Jubert Hermoso, Mo~Metanat, Mohammad Rastegari, Munish Bansal, Nandhini Santhanam, Natascha Parks, Natasha White, Navyata Bawa, Nayan Singhal, Nick Egebo, Nicolas Usunier, Nikhil Mehta, Nikolay~Pavlovich Laptev, Ning Dong, Norman Cheng, Oleg Chernoguz, Olivia Hart, Omkar Salpekar, Ozlem Kalinli, Parkin Kent, Parth Parekh, Paul Saab, Pavan Balaji, Pedro Rittner, Philip Bontrager, Pierre Roux, Piotr Dollar, Polina Zvyagina, Prashant Ratanchandani, Pritish Yuvraj, Qian Liang, Rachad Alao, Rachel Rodriguez, Rafi Ayub, Raghotham Murthy, Raghu Nayani, Rahul Mitra, Rangaprabhu Parthasarathy, Raymond Li, Rebekkah Hogan, Robin Battey, Rocky Wang, Russ Howes, Ruty Rinott, Sachin Mehta, Sachin Siby, Sai~Jayesh Bondu, Samyak Datta, Sara Chugh, Sara Hunt, Sargun Dhillon, Sasha Sidorov, Satadru Pan, Saurabh Mahajan,
  Saurabh Verma, Seiji Yamamoto, Sharadh Ramaswamy, Shaun Lindsay, Shaun Lindsay, Sheng Feng, Shenghao Lin, Shengxin~Cindy Zha, Shishir Patil, Shiva Shankar, Shuqiang Zhang, Shuqiang Zhang, Sinong Wang, Sneha Agarwal, Soji Sajuyigbe, Soumith Chintala, Stephanie Max, Stephen Chen, Steve Kehoe, Steve Satterfield, Sudarshan Govindaprasad, Sumit Gupta, Summer Deng, Sungmin Cho, Sunny Virk, Suraj Subramanian, Sy~Choudhury, Sydney Goldman, Tal Remez, Tamar Glaser, Tamara Best, Thilo Koehler, Thomas Robinson, Tianhe Li, Tianjun Zhang, Tim Matthews, Timothy Chou, Tzook Shaked, Varun Vontimitta, Victoria Ajayi, Victoria Montanez, Vijai Mohan, Vinay~Satish Kumar, Vishal Mangla, Vlad Ionescu, Vlad Poenaru, Vlad~Tiberiu Mihailescu, Vladimir Ivanov, Wei Li, Wenchen Wang, Wenwen Jiang, Wes Bouaziz, Will Constable, Xiaocheng Tang, Xiaojian Wu, Xiaolan Wang, Xilun Wu, Xinbo Gao, Yaniv Kleinman, Yanjun Chen, Ye~Hu, Ye~Jia, Ye~Qi, Yenda Li, Yilin Zhang, Ying Zhang, Yossi Adi, Youngjin Nam, Yu, Wang, Yu~Zhao, Yuchen Hao, Yundi
  Qian, Yunlu Li, Yuzi He, Zach Rait, Zachary DeVito, Zef Rosnbrick, Zhaoduo Wen, Zhenyu Yang, Zhiwei Zhao, and Zhiyu Ma.
\newblock The {Llama} 3 {Herd} of {Models}, November 2024.
\newblock arXiv:2407.21783 [cs].

\bibitem{griot_large_2025}
Maxime Griot, Coralie Hemptinne, Jean Vanderdonckt, and Demet Yuksel.
\newblock Large {Language} {Models} lack essential metacognition for reliable medical reasoning.
\newblock {\em Nature Communications}, 16(1):642, January 2025.
\newblock Publisher: Nature Publishing Group.

\bibitem{gu_survey_2025}
Jiawei Gu, Xuhui Jiang, Zhichao Shi, Hexiang Tan, Xuehao Zhai, Chengjin Xu, Wei Li, Yinghan Shen, Shengjie Ma, Honghao Liu, Saizhuo Wang, Kun Zhang, Yuanzhuo Wang, Wen Gao, Lionel Ni, and Jian Guo.
\newblock A {Survey} on {LLM}-as-a-{Judge}, March 2025.
\newblock arXiv:2411.15594 [cs].

\bibitem{hao_retrospective_2025}
Yuexing Hao, Jason Holmes, Jared Hobson, Alexandra Bennett, Elizabeth~L. McKone, Daniel~K. Ebner, David~M. Routman, Satomi Shiraishi, Samir~H. Patel, Nathan~Y. Yu, Chris~L. Hallemeier, Brooke~E. Ball, Mark Waddle, and Wei Liu.
\newblock Retrospective {Comparative} {Analysis} of {Prostate} {Cancer} {In}-{Basket} {Messages}: {Responses} {From} {Closed}-{Domain} {Large} {Language} {Models} {Versus} {Clinical} {Teams}.
\newblock {\em Mayo Clinic Proceedings: Digital Health}, 3(1):100198, March 2025.

\bibitem{hendrycks_measuring_2021}
Dan Hendrycks, Collin Burns, Steven Basart, Andy Zou, Mantas Mazeika, Dawn Song, and Jacob Steinhardt.
\newblock Measuring {Massive} {Multitask} {Language} {Understanding}, January 2021.
\newblock arXiv:2009.03300 [cs].

\bibitem{ishii_analysis_2024}
Ai~Ishii, Naoya Inoue, Hisami Suzuki, and Satoshi Sekine.
\newblock Analysis of {LLM}`s “{Spurious}” {Correct} {Answers} {Using} {Evidence} {Information} of {Multi}-hop {QA} {Datasets}.
\newblock In Russa Biswas, Lucie-Aimée Kaffee, Oshin Agarwal, Pasquale Minervini, Sameer Singh, and Gerard de~Melo, editors, {\em Proceedings of the 1st {Workshop} on {Knowledge} {Graphs} and {Large} {Language} {Models} ({KaLLM} 2024)}, pages 24--34, Bangkok, Thailand, August 2024. Association for Computational Linguistics.

\bibitem{jin_rjua-meddqa_2024}
Congyun Jin, Ming Zhang, Weixiao Ma, Yujiao Li, Yingbo Wang, Yabo Jia, Yuliang Du, Tao Sun, Haowen Wang, Cong Fan, Jinjie Gu, Chenfei Chi, Xiangguo Lv, Fangzhou Li, Wei Xue, and Yiran Huang.
\newblock {RJUA}-{MedDQA}: {A} {Multimodal} {Benchmark} for {Medical} {Document} {Question} {Answering} and {Clinical} {Reasoning}.
\newblock In {\em Proceedings of the 30th {ACM} {SIGKDD} {Conference} on {Knowledge} {Discovery} and {Data} {Mining}}, {KDD} '24, pages 5218--5229, New York, NY, USA, August 2024. Association for Computing Machinery.

\bibitem{jin_what_2021}
Di~Jin, Eileen Pan, Nassim Oufattole, Wei-Hung Weng, Hanyi Fang, and Peter Szolovits.
\newblock What {Disease} {Does} {This} {Patient} {Have}? {A} {Large}-{Scale} {Open} {Domain} {Question} {Answering} {Dataset} from {Medical} {Exams}.
\newblock {\em Applied Sciences}, 11(14):6421, January 2021.
\newblock Number: 14 Publisher: Multidisciplinary Digital Publishing Institute.

\bibitem{jin_pubmedqa_2019}
Qiao Jin, Bhuwan Dhingra, Zhengping Liu, William Cohen, and Xinghua Lu.
\newblock {PubMedQA}: {A} {Dataset} for {Biomedical} {Research} {Question} {Answering}.
\newblock In Kentaro Inui, Jing Jiang, Vincent Ng, and Xiaojun Wan, editors, {\em Proceedings of the 2019 {Conference} on {Empirical} {Methods} in {Natural} {Language} {Processing} and the 9th {International} {Joint} {Conference} on {Natural} {Language} {Processing} ({EMNLP}-{IJCNLP})}, pages 2567--2577, Hong Kong, China, November 2019. Association for Computational Linguistics.

\bibitem{kabongo_effective_2024}
Salomon Kabongo, Jennifer D’Souza, and Sören Auer.
\newblock Effective {Context} {Selection} in {LLM}-{Based} {Leaderboard} {Generation}: {An} {Empirical} {Study}.
\newblock In Amon Rapp, Luigi Di~Caro, Farid Meziane, and Vijayan Sugumaran, editors, {\em Natural {Language} {Processing} and {Information} {Systems}}, pages 150--160, Cham, 2024. Springer Nature Switzerland.

\bibitem{katz_gpt_2024}
Uriel Katz, Eran Cohen, Eliya Shachar, Jonathan Somer, Adam Fink, Eli Morse, Beki Shreiber, and Ido Wolf.
\newblock {GPT} versus {Resident} {Physicians} — {A} {Benchmark} {Based} on {Official} {Board} {Scores}.
\newblock {\em NEJM AI}, 1(5):AIdbp2300192, April 2024.
\newblock Publisher: Massachusetts Medical Society.

\bibitem{khattab_baleen_2021}
Omar Khattab, Christopher Potts, and Matei Zaharia.
\newblock Baleen: robust multi-hop reasoning at scale via condensed retrieval.
\newblock In {\em Proceedings of the 35th {International} {Conference} on {Neural} {Information} {Processing} {Systems}}, {NIPS} '21, pages 27670--27682, Red Hook, NY, USA, December 2021. Curran Associates Inc.

\bibitem{li_mediq_2024}
Shuyue~S. Li, Vidhisha Balachandran, Shangbin Feng, Jonathan~S. Ilgen, Emma Pierson, Pang~W. Koh, and Yulia Tsvetkov.
\newblock {MediQ}: {Question}-{Asking} {LLMs} and a {Benchmark} for {Reliable} {Interactive} {Clinical} {Reasoning}.
\newblock {\em Advances in Neural Information Processing Systems}, 37:28858--28888, December 2024.

\bibitem{liu_attribot_2024}
Fengyuan Liu, Nikhil Kandpal, and Colin Raffel.
\newblock {AttriBoT}: {A} {Bag} of {Tricks} for {Efficiently} {Approximating} {Leave}-{One}-{Out} {Context} {Attribution}.
\newblock October 2024.

\bibitem{lu_learn_2022}
Pan Lu, Swaroop Mishra, Tony Xia, Liang Qiu, Kai-Wei Chang, Song-Chun Zhu, Oyvind Tafjord, Peter Clark, and Ashwin Kalyan.
\newblock Learn to {Explain}: {Multimodal} {Reasoning} via {Thought} {Chains} for {Science} {Question} {Answering}.
\newblock October 2022.

\bibitem{mcduff_towards_2025}
Daniel McDuff, Mike Schaekermann, Tao Tu, Anil Palepu, Amy Wang, Jake Garrison, Karan Singhal, Yash Sharma, Shekoofeh Azizi, Kavita Kulkarni, Le~Hou, Yong Cheng, Yun Liu, S.~Sara Mahdavi, Sushant Prakash, Anupam Pathak, Christopher Semturs, Shwetak Patel, Dale~R. Webster, Ewa Dominowska, Juraj Gottweis, Joelle Barral, Katherine Chou, Greg~S. Corrado, Yossi Matias, Jake Sunshine, Alan Karthikesalingam, and Vivek Natarajan.
\newblock Towards accurate differential diagnosis with large language models.
\newblock {\em Nature}, pages 1--7, April 2025.
\newblock Publisher: Nature Publishing Group.

\bibitem{mckechnie_context_2025}
Jack McKechnie, Graham McDonald, and Craig Macdonald.
\newblock Context {Example} {Selection} for {LLM} {Generated} {Relevance} {Assessments}.
\newblock In {\em Advances in {Information} {Retrieval}: 47th {European} {Conference} on {Information} {Retrieval}, {ECIR} 2025, {Lucca}, {Italy}, {April} 6–10, 2025, {Proceedings}, {Part} {I}}, pages 293--309, Berlin, Heidelberg, April 2025. Springer-Verlag.

\bibitem{openai_gpt-4o_2024}
OpenAI, Aaron Hurst, Adam Lerer, Adam~P. Goucher, Adam Perelman, Aditya Ramesh, Aidan Clark, A.~J. Ostrow, Akila Welihinda, Alan Hayes, Alec Radford, Aleksander Mądry, Alex Baker-Whitcomb, Alex Beutel, Alex Borzunov, Alex Carney, Alex Chow, Alex Kirillov, Alex Nichol, Alex Paino, Alex Renzin, Alex~Tachard Passos, Alexander Kirillov, Alexi Christakis, Alexis Conneau, Ali Kamali, Allan Jabri, Allison Moyer, Allison Tam, Amadou Crookes, Amin Tootoochian, Amin Tootoonchian, Ananya Kumar, Andrea Vallone, Andrej Karpathy, Andrew Braunstein, Andrew Cann, Andrew Codispoti, Andrew Galu, Andrew Kondrich, Andrew Tulloch, Andrey Mishchenko, Angela Baek, Angela Jiang, Antoine Pelisse, Antonia Woodford, Anuj Gosalia, Arka Dhar, Ashley Pantuliano, Avi Nayak, Avital Oliver, Barret Zoph, Behrooz Ghorbani, Ben Leimberger, Ben Rossen, Ben Sokolowsky, Ben Wang, Benjamin Zweig, Beth Hoover, Blake Samic, Bob McGrew, Bobby Spero, Bogo Giertler, Bowen Cheng, Brad Lightcap, Brandon Walkin, Brendan Quinn, Brian Guarraci, Brian Hsu,
  Bright Kellogg, Brydon Eastman, Camillo Lugaresi, Carroll Wainwright, Cary Bassin, Cary Hudson, Casey Chu, Chad Nelson, Chak Li, Chan~Jun Shern, Channing Conger, Charlotte Barette, Chelsea Voss, Chen Ding, Cheng Lu, Chong Zhang, Chris Beaumont, Chris Hallacy, Chris Koch, Christian Gibson, Christina Kim, Christine Choi, Christine McLeavey, Christopher Hesse, Claudia Fischer, Clemens Winter, Coley Czarnecki, Colin Jarvis, Colin Wei, Constantin Koumouzelis, Dane Sherburn, Daniel Kappler, Daniel Levin, Daniel Levy, David Carr, David Farhi, David Mely, David Robinson, David Sasaki, Denny Jin, Dev Valladares, Dimitris Tsipras, Doug Li, Duc~Phong Nguyen, Duncan Findlay, Edede Oiwoh, Edmund Wong, Ehsan Asdar, Elizabeth Proehl, Elizabeth Yang, Eric Antonow, Eric Kramer, Eric Peterson, Eric Sigler, Eric Wallace, Eugene Brevdo, Evan Mays, Farzad Khorasani, Felipe~Petroski Such, Filippo Raso, Francis Zhang, Fred~von Lohmann, Freddie Sulit, Gabriel Goh, Gene Oden, Geoff Salmon, Giulio Starace, Greg Brockman, Hadi
  Salman, Haiming Bao, Haitang Hu, Hannah Wong, Haoyu Wang, Heather Schmidt, Heather Whitney, Heewoo Jun, Hendrik Kirchner, Henrique Ponde de~Oliveira Pinto, Hongyu Ren, Huiwen Chang, Hyung~Won Chung, Ian Kivlichan, Ian O'Connell, Ian O'Connell, Ian Osband, Ian Silber, Ian Sohl, Ibrahim Okuyucu, Ikai Lan, Ilya Kostrikov, Ilya Sutskever, Ingmar Kanitscheider, Ishaan Gulrajani, Jacob Coxon, Jacob Menick, Jakub Pachocki, James Aung, James Betker, James Crooks, James Lennon, Jamie Kiros, Jan Leike, Jane Park, Jason Kwon, Jason Phang, Jason Teplitz, Jason Wei, Jason Wolfe, Jay Chen, Jeff Harris, Jenia Varavva, Jessica~Gan Lee, Jessica Shieh, Ji~Lin, Jiahui Yu, Jiayi Weng, Jie Tang, Jieqi Yu, Joanne Jang, Joaquin~Quinonero Candela, Joe Beutler, Joe Landers, Joel Parish, Johannes Heidecke, John Schulman, Jonathan Lachman, Jonathan McKay, Jonathan Uesato, Jonathan Ward, Jong~Wook Kim, Joost Huizinga, Jordan Sitkin, Jos Kraaijeveld, Josh Gross, Josh Kaplan, Josh Snyder, Joshua Achiam, Joy Jiao, Joyce Lee, Juntang
  Zhuang, Justyn Harriman, Kai Fricke, Kai Hayashi, Karan Singhal, Katy Shi, Kavin Karthik, Kayla Wood, Kendra Rimbach, Kenny Hsu, Kenny Nguyen, Keren Gu-Lemberg, Kevin Button, Kevin Liu, Kiel Howe, Krithika Muthukumar, Kyle Luther, Lama Ahmad, Larry Kai, Lauren Itow, Lauren Workman, Leher Pathak, Leo Chen, Li~Jing, Lia Guy, Liam Fedus, Liang Zhou, Lien Mamitsuka, Lilian Weng, Lindsay McCallum, Lindsey Held, Long Ouyang, Louis Feuvrier, Lu~Zhang, Lukas Kondraciuk, Lukasz Kaiser, Luke Hewitt, Luke Metz, Lyric Doshi, Mada Aflak, Maddie Simens, Madelaine Boyd, Madeleine Thompson, Marat Dukhan, Mark Chen, Mark Gray, Mark Hudnall, Marvin Zhang, Marwan Aljubeh, Mateusz Litwin, Matthew Zeng, Max Johnson, Maya Shetty, Mayank Gupta, Meghan Shah, Mehmet Yatbaz, Meng~Jia Yang, Mengchao Zhong, Mia Glaese, Mianna Chen, Michael Janner, Michael Lampe, Michael Petrov, Michael Wu, Michele Wang, Michelle Fradin, Michelle Pokrass, Miguel Castro, Miguel Oom Temudo~de Castro, Mikhail Pavlov, Miles Brundage, Miles Wang, Minal
  Khan, Mira Murati, Mo~Bavarian, Molly Lin, Murat Yesildal, Nacho Soto, Natalia Gimelshein, Natalie Cone, Natalie Staudacher, Natalie Summers, Natan LaFontaine, Neil Chowdhury, Nick Ryder, Nick Stathas, Nick Turley, Nik Tezak, Niko Felix, Nithanth Kudige, Nitish Keskar, Noah Deutsch, Noel Bundick, Nora Puckett, Ofir Nachum, Ola Okelola, Oleg Boiko, Oleg Murk, Oliver Jaffe, Olivia Watkins, Olivier Godement, Owen Campbell-Moore, Patrick Chao, Paul McMillan, Pavel Belov, Peng Su, Peter Bak, Peter Bakkum, Peter Deng, Peter Dolan, Peter Hoeschele, Peter Welinder, Phil Tillet, Philip Pronin, Philippe Tillet, Prafulla Dhariwal, Qiming Yuan, Rachel Dias, Rachel Lim, Rahul Arora, Rajan Troll, Randall Lin, Rapha~Gontijo Lopes, Raul Puri, Reah Miyara, Reimar Leike, Renaud Gaubert, Reza Zamani, Ricky Wang, Rob Donnelly, Rob Honsby, Rocky Smith, Rohan Sahai, Rohit Ramchandani, Romain Huet, Rory Carmichael, Rowan Zellers, Roy Chen, Ruby Chen, Ruslan Nigmatullin, Ryan Cheu, Saachi Jain, Sam Altman, Sam Schoenholz, Sam
  Toizer, Samuel Miserendino, Sandhini Agarwal, Sara Culver, Scott Ethersmith, Scott Gray, Sean Grove, Sean Metzger, Shamez Hermani, Shantanu Jain, Shengjia Zhao, Sherwin Wu, Shino Jomoto, Shirong Wu, Shuaiqi, Xia, Sonia Phene, Spencer Papay, Srinivas Narayanan, Steve Coffey, Steve Lee, Stewart Hall, Suchir Balaji, Tal Broda, Tal Stramer, Tao Xu, Tarun Gogineni, Taya Christianson, Ted Sanders, Tejal Patwardhan, Thomas Cunninghman, Thomas Degry, Thomas Dimson, Thomas Raoux, Thomas Shadwell, Tianhao Zheng, Todd Underwood, Todor Markov, Toki Sherbakov, Tom Rubin, Tom Stasi, Tomer Kaftan, Tristan Heywood, Troy Peterson, Tyce Walters, Tyna Eloundou, Valerie Qi, Veit Moeller, Vinnie Monaco, Vishal Kuo, Vlad Fomenko, Wayne Chang, Weiyi Zheng, Wenda Zhou, Wesam Manassra, Will Sheu, Wojciech Zaremba, Yash Patil, Yilei Qian, Yongjik Kim, Youlong Cheng, Yu~Zhang, Yuchen He, Yuchen Zhang, Yujia Jin, Yunxing Dai, and Yury Malkov.
\newblock {GPT}-4o {System} {Card}, October 2024.
\newblock arXiv:2410.21276 [cs].

\bibitem{pal_medmcqa_2022}
Ankit Pal, Logesh~Kumar Umapathi, and Malaikannan Sankarasubbu.
\newblock {MedMCQA}: {A} {Large}-scale {Multi}-{Subject} {Multi}-{Choice} {Dataset} for {Medical} domain {Question} {Answering}.
\newblock In {\em Proceedings of the {Conference} on {Health}, {Inference}, and {Learning}}, pages 248--260. PMLR, April 2022.
\newblock ISSN: 2640-3498.

\bibitem{panickssery_llm_2024}
Arjun Panickssery, Samuel~R. Bowman, and Shi Feng.
\newblock {LLM} {Evaluators} {Recognize} and {Favor} {Their} {Own} {Generations}.
\newblock November 2024.

\bibitem{park_does_2015}
Wan~Beom Park, Seok~Hoon Kang, Yoon-Seong Lee, and Sun~Jung Myung.
\newblock Does {Objective} {Structured} {Clinical} {Examinations} {Score} {Reflect} the {Clinical} {Reasoning} {Ability} of {Medical} {Students}?
\newblock {\em The American Journal of the Medical Sciences}, 350(1):64--67, July 2015.

\bibitem{qwen_qwen25_2025}
Qwen, An~Yang, Baosong Yang, Beichen Zhang, Binyuan Hui, Bo~Zheng, Bowen Yu, Chengyuan Li, Dayiheng Liu, Fei Huang, Haoran Wei, Huan Lin, Jian Yang, Jianhong Tu, Jianwei Zhang, Jianxin Yang, Jiaxi Yang, Jingren Zhou, Junyang Lin, Kai Dang, Keming Lu, Keqin Bao, Kexin Yang, Le~Yu, Mei Li, Mingfeng Xue, Pei Zhang, Qin Zhu, Rui Men, Runji Lin, Tianhao Li, Tianyi Tang, Tingyu Xia, Xingzhang Ren, Xuancheng Ren, Yang Fan, Yang Su, Yichang Zhang, Yu~Wan, Yuqiong Liu, Zeyu Cui, Zhenru Zhang, and Zihan Qiu.
\newblock Qwen2.5 {Technical} {Report}, January 2025.
\newblock arXiv:2412.15115 [cs].

\bibitem{razavi_benchmarking_2025}
Amirhossein Razavi, Mina Soltangheis, Negar Arabzadeh, Sara Salamat, Morteza Zihayat, and Ebrahim Bagheri.
\newblock Benchmarking {Prompt} {Sensitivity} in {Large} {Language} {Models}.
\newblock In {\em Advances in {Information} {Retrieval}: 47th {European} {Conference} on {Information} {Retrieval}, {ECIR} 2025, {Lucca}, {Italy}, {April} 6–10, 2025, {Proceedings}, {Part} {III}}, pages 303--313, Berlin, Heidelberg, April 2025. Springer-Verlag.

\bibitem{rong_towards_2024}
Yao Rong, Tobias Leemann, Thai-Trang Nguyen, Lisa Fiedler, Peizhu Qian, Vaibhav Unhelkar, Tina Seidel, Gjergji Kasneci, and Enkelejda Kasneci.
\newblock Towards {Human}-{Centered} {Explainable} {AI}: {A} {Survey} of {User} {Studies} for {Model} {Explanations}.
\newblock {\em IEEE Trans. Pattern Anal. Mach. Intell.}, 46(4):2104--2122, April 2024.

\bibitem{savage_diagnostic_2024}
Thomas Savage, Ashwin Nayak, Robert Gallo, Ekanath Rangan, and Jonathan~H. Chen.
\newblock Diagnostic reasoning prompts reveal the potential for large language model interpretability in medicine.
\newblock {\em npj Digital Medicine}, 7(1):1--7, January 2024.
\newblock Publisher: Nature Publishing Group.

\bibitem{schuler_context_2025}
Katharina Schuler, Ian-C. Jung, Maria Zerlik, Waldemar Hahn, Martin Sedlmayr, and Brita Sedlmayr.
\newblock Context factors in clinical decision-making: a scoping review.
\newblock {\em BMC Medical Informatics and Decision Making}, 25(1):133, March 2025.

\bibitem{sharma_relevance_2017}
Shikhar Sharma, Layla~El Asri, Hannes Schulz, and Jeremie Zumer.
\newblock Relevance of {Unsupervised} {Metrics} in {Task}-{Oriented} {Dialogue} for {Evaluating} {Natural} {Language} {Generation}, June 2017.
\newblock arXiv:1706.09799 [cs].

\bibitem{shi_judging_2025}
Lin Shi, Chiyu Ma, Wenhua Liang, Xingjian Diao, Weicheng Ma, and Soroush Vosoughi.
\newblock Judging the {Judges}: {A} {Systematic} {Study} of {Position} {Bias} in {LLM}-as-a-{Judge}, April 2025.
\newblock arXiv:2406.07791 [cs].

\bibitem{singhal_toward_2025}
Karan Singhal, Tao Tu, Juraj Gottweis, Rory Sayres, Ellery Wulczyn, Mohamed Amin, Le~Hou, Kevin Clark, Stephen~R. Pfohl, Heather Cole-Lewis, Darlene Neal, Qazi~Mamunur Rashid, Mike Schaekermann, Amy Wang, Dev Dash, Jonathan~H. Chen, Nigam~H. Shah, Sami Lachgar, Philip~Andrew Mansfield, Sushant Prakash, Bradley Green, Ewa Dominowska, Blaise Agüera~y Arcas, Nenad Tomašev, Yun Liu, Renee Wong, Christopher Semturs, S.~Sara Mahdavi, Joelle~K. Barral, Dale~R. Webster, Greg~S. Corrado, Yossi Matias, Shekoofeh Azizi, Alan Karthikesalingam, and Vivek Natarajan.
\newblock Toward expert-level medical question answering with large language models.
\newblock {\em Nature Medicine}, 31(3):943--950, March 2025.
\newblock Publisher: Nature Publishing Group.

\bibitem{soboroff_dont_2025}
Ian Soboroff.
\newblock Don't {Use} {LLMs} to {Make} {Relevance} {Judgments}.
\newblock {\em Information Retrieval Research}, 1(1):29--46, March 2025.
\newblock Number: 1.

\bibitem{soni_radqa_2022}
Sarvesh Soni, Meghana Gudala, Atieh Pajouhi, and Kirk Roberts.
\newblock {RadQA}: {A} {Question} {Answering} {Dataset} to {Improve} {Comprehension} of {Radiology} {Reports}.
\newblock In Nicoletta Calzolari, Frédéric Béchet, Philippe Blache, Khalid Choukri, Christopher Cieri, Thierry Declerck, Sara Goggi, Hitoshi Isahara, Bente Maegaard, Joseph Mariani, Hélène Mazo, Jan Odijk, and Stelios Piperidis, editors, {\em Proceedings of the {Thirteenth} {Language} {Resources} and {Evaluation} {Conference}}, pages 6250--6259, Marseille, France, June 2022. European Language Resources Association.

\bibitem{strong_chatbot_2023}
Eric Strong, Alicia DiGiammarino, Yingjie Weng, Andre Kumar, Poonam Hosamani, Jason Hom, and Jonathan~H. Chen.
\newblock Chatbot vs {Medical} {Student} {Performance} on {Free}-{Response} {Clinical} {Reasoning} {Examinations}.
\newblock {\em JAMA internal medicine}, 183(9):1028--1030, September 2023.

\bibitem{toma_clinical_2023}
Augustin Toma, Patrick~R. Lawler, Jimmy Ba, Rahul~G. Krishnan, Barry~B. Rubin, and Bo~Wang.
\newblock Clinical {Camel}: {An} {Open} {Expert}-{Level} {Medical} {Language} {Model} with {Dialogue}-{Based} {Knowledge} {Encoding}, August 2023.
\newblock arXiv:2305.12031 [cs].

\bibitem{wang_prompt_2024}
Li~Wang, Xi~Chen, XiangWen Deng, Hao Wen, MingKe You, WeiZhi Liu, Qi~Li, and Jian Li.
\newblock Prompt engineering in consistency and reliability with the evidence-based guideline for {LLMs}.
\newblock {\em npj Digital Medicine}, 7(1):1--9, February 2024.
\newblock Publisher: Nature Publishing Group.

\bibitem{wu_medreason_2025}
Juncheng Wu, Wenlong Deng, Xingxuan Li, Sheng Liu, Taomian Mi, Yifan Peng, Ziyang Xu, Yi~Liu, Hyunjin Cho, Chang-In Choi, Yihan Cao, Hui Ren, Xiang Li, Xiaoxiao Li, and Yuyin Zhou.
\newblock {MedReason}: {Eliciting} {Factual} {Medical} {Reasoning} {Steps} in {LLMs} via {Knowledge} {Graphs}, April 2025.
\newblock arXiv:2504.00993 [cs].

\bibitem{wu_automated_2025}
Kevin Wu, Eric Wu, Kevin Wei, Angela Zhang, Allison Casasola, Teresa Nguyen, Sith Riantawan, Patricia Shi, Daniel Ho, and James Zou.
\newblock An automated framework for assessing how well {LLMs} cite relevant medical references.
\newblock {\em Nature Communications}, 16(1):3615, April 2025.
\newblock Publisher: Nature Publishing Group.

\bibitem{xia_cares_2024}
Peng Xia, Ze~Chen, Juanxi Tian, Yangrui Gong, Ruibo Hou, Yue Xu, Zhenbang Wu, Zhiyuan Fan, Yiyang Zhou, Kangyu Zhu, Wenhao Zheng, Zhaoyang Wang, Xiao Wang, Xuchao Zhang, Chetan Bansal, Marc Niethammer, Junzhou Huang, Hongtu Zhu, Yun Li, Jimeng Sun, Zongyuan Ge, Gang Li, James Zou, and Huaxiu Yao.
\newblock {CARES}: {A} {Comprehensive} {Benchmark} of {Trustworthiness} in {Medical} {Vision} {Language} {Models}.
\newblock {\em Advances in Neural Information Processing Systems}, 37:140334--140365, December 2024.

\bibitem{yang_harnessing_2023}
Qian Yang, Yuexing Hao, Kexin Quan, Stephen Yang, Yiran Zhao, Volodymyr Kuleshov, and Fei Wang.
\newblock Harnessing {Biomedical} {Literature} to {Calibrate} {Clinicians}’ {Trust} in {AI} {Decision} {Support} {Systems}.
\newblock In {\em Proceedings of the 2023 {CHI} {Conference} on {Human} {Factors} in {Computing} {Systems}}, {CHI} '23, pages 1--14, New York, NY, USA, April 2023. Association for Computing Machinery.

\bibitem{zhang_survey_2024}
Deshiwei Zhang, Xiaojuan Xue, Peng Gao, Zhijuan Jin, Menghan Hu, Yue Wu, and Xiayang Ying.
\newblock A survey of datasets in medicine for large language models.
\newblock {\em Intelligence \& Robotics}, 4(4):457--478, December 2024.
\newblock Publisher: OAE Publishing Inc.

\bibitem{zhao_moverscore_2019}
Wei Zhao, Maxime Peyrard, Fei Liu, Yang Gao, Christian~M. Meyer, and Steffen Eger.
\newblock {MoverScore}: {Text} {Generation} {Evaluating} with {Contextualized} {Embeddings} and {Earth} {Mover} {Distance}.
\newblock In Kentaro Inui, Jing Jiang, Vincent Ng, and Xiaojun Wan, editors, {\em Proceedings of the 2019 {Conference} on {Empirical} {Methods} in {Natural} {Language} {Processing} and the 9th {International} {Joint} {Conference} on {Natural} {Language} {Processing} ({EMNLP}-{IJCNLP})}, pages 563--578, Hong Kong, China, November 2019. Association for Computational Linguistics.

\bibitem{zheng_judging_2023}
Lianmin Zheng, Wei-Lin Chiang, Ying Sheng, Siyuan Zhuang, Zhanghao Wu, Yonghao Zhuang, Zi~Lin, Zhuohan Li, Dacheng Li, Eric Xing, Hao Zhang, Joseph~E. Gonzalez, and Ion Stoica.
\newblock Judging {LLM}-as-a-{Judge} with {MT}-{Bench} and {Chatbot} {Arena}.
\newblock {\em Advances in Neural Information Processing Systems}, 36:46595--46623, December 2023.

\bibitem{zhou_explainable_2025}
Shuang Zhou, Mingquan Lin, Sirui Ding, Jiashuo Wang, Canyu Chen, Genevieve~B. Melton, James Zou, and Rui Zhang.
\newblock Explainable differential diagnosis with dual-inference large language models.
\newblock {\em npj Health Systems}, 2(1):1--9, April 2025.
\newblock Publisher: Nature Publishing Group.

\bibitem{zuo_medxpertqa_2025}
Yuxin Zuo, Shang Qu, Yifei Li, Zhangren Chen, Xuekai Zhu, Ermo Hua, Kaiyan Zhang, Ning Ding, and Bowen Zhou.
\newblock {MedXpertQA}: {Benchmarking} {Expert}-{Level} {Medical} {Reasoning} and {Understanding}, February 2025.
\newblock arXiv:2501.18362 [cs].

\end{thebibliography}

\newpage
\appendix

\section{Dataset Explanation}
\label{dataset_explanation}
\textbf{Massive Multitask Language Understanding (MMLU)} is a common benchmark which consists of multiple domains and tasks based on real-world exams \cite{hendrycks_measuring_2021}. It includes 57 subjects across STEM, the humanities, the social sciences. Here we only focused on medical related questions (precision\_medicine), which has 272 multiple-choice medical questions. 

The \textbf{JAMA Clinical Challenge dataset} includes 1,034 clinical cases sourced from the JAMA Network Clinical Challenge archive. Each entry summarizes a real and diagnostically complex clinical scenario, presented in the form of a question. These challenges feature an extended case vignette followed by a multiple-choice question with four answer options, accompanied by a detailed discussion explaining both the correct and incorrect responses. The questions span a broad spectrum of medical topics \cite{chen_benchmarking_2025}.

\textbf{Medbullets} consists of 298 United States Medical Licensing Examination (USMLE) Step 2 and Step 3–style questions curated from open-access posts beginning in April 2022. These questions aim to reflect common clinical scenarios encountered in medical education, with difficulty levels comparable to Step 2 and 3 exams. Each item includes a brief case description, five answer choices, and an explanation that clarifies the reasoning behind both correct and incorrect responses. Compared to JAMA, these cases tend to be shorter and potentially less complex \cite{chen_benchmarking_2025}.

\textbf{MedXpertQA} consists of 2450 questions for text evaluation. It is a highly challenging
and comprehensive medical multiple-choice benchmark. MedXpertQA integrates specialty-specific assessments into medical benchmarking and challenging medical exam questions with real-world clinical information into medical multimodal benchmarking \cite{zuo_medxpertqa_2025}.

\subsection{NLP Analysis}
\begin{figure}[h!]
  \centering
  \includegraphics[width=\textwidth]{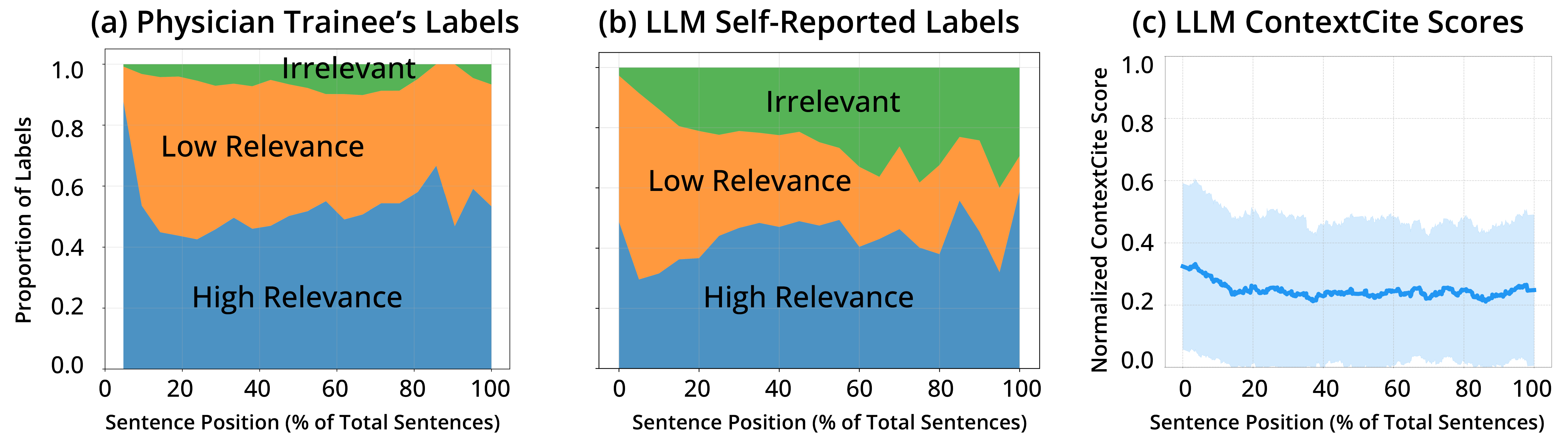}
  \caption{\textbf{Sentence Position Analysis.} Plot (a) Distribution of physician trainees’ majority‐vote relevance labels by sentence position. Plot (b) Distribution of GPT-4o self-reported relevance labels by sentence position. Plot (c) ContextCite scores across the context for three open-source models (Qwen-14B, Llama-70B, Qwen-72B).}
  \label{fig:sentence_position}
\end{figure}

In order to investigate how sentence relevance shifts according to its position in the clinical vignette, we plotted the labels assigned by physician trainees and those self reported by LLMs (Figure \ref{fig:sentence_position} plots (a), (b)) and LLM ContextCite scores (Figure \ref{fig:sentence_position} plot (c)). Our objective was to determine whether trainees or the model demonstrate systematic attention to particular segments of the patient profile. All three plots indicate that sentences appearing at the beginning of the text receive the highest relevance ratings. GPT-4o marks slightly fewer sentences as highly relevant and more as irrelevant in the central region compared with physician trainees. In contrast, ContextCite scores decline from approximately 0.33 at the outset to 0.22 by the tenth percentile, then plateau between 0.20 and 0.25 with minimal variance. This flat, low-variance profile diverges sharply from the dynamic patterns of expert and self-reported labels, suggesting that ContextCite does not capture the nuanced, position-dependent relevance judgments.

% --- Figure 2: Attribution Score Comparisons ---

% \begin{wrapfigure}{r}{0.65\textwidth}
%   \centering
%   \includegraphics[width=\linewidth]{Figures/CC_SR_Removal.pdf}
%   \caption{\textbf{Physician trainee relevance label distribution across sentence positions in four medical QA datasets.}}
%   \label{Self_Report_Vs_Centaur_Lab}
% \end{wrapfigure}

\section{Expert Annotation Dataset Interpretation}

We instructed each labeler to annotate every sentence as “high relevance,” “low relevance,” or “irrelevant.” Depending on question difficulty and the exclusion of annotations from labelers whose answer classifications proved incorrect, each item received between one and three valid annotations. We then assigned numeric scores to those labels: high relevance was scored as 1.0, low relevance as 0.5, and irrelevant as 0.0. For sentences with more than one annotation, we calculated the average of those scores. If the average exceeded 0.66, we classified the sentence as high relevance; if it fell between 0.33 and 0.66, we classified it as low relevance; and if it was below 0.33, we classified it as irrelevant. The specific rules and combinations of relevance labels are displayed in Table \ref{tab:label_landscape_comparison}.

\begin{table}[h!]
\centering
\begin{tabular}{l|c|c|c}
\toprule
\textbf{Labelers} & \makecell[c]{High Relevance \\ Labels} & \makecell[c]{Low Relevance \\ Labels} & \makecell[c]{Irrelevant Labels}\\
\midrule
3 Correct Labels & \makecell[c]{High, High, High \\ High, High, Low \\ High, High, Irr \\ High, Low, Low  } & \makecell[c]{Low, Low, Low \\ High, Low, Irr } & \makecell[c]{High, Irr, Irr\\ Low, Low, Irr \\ Low, Irr, Irr \\ Irr, Irr, Irr } \\
\hline
2 Correct Labels & \makecell[c]{High, High \\ High, Low} & \makecell[c]{High, Irr \\ Low, Low  \\ Low, Irr}  & Irr, Irr \\
\hline
1 Correct Labels & High & Low & Irr \\
\bottomrule
\end{tabular}
\vspace{0.5em}

\noindent* “High” refers to high relevance; “Low” to low relevance; “Irr” to irrelevant.

\vspace{0.5em}
\caption{Rules based on majority label agreement across different label landscapes for each sentence-level analysis.}
\label{tab:label_landscape_comparison}
\end{table}

To structure the evaluation, we designed a framework distinguishing between accurate prediction and relevance agreement, summarized in the confusion matrix presented in Table \ref{Confusion_Matrix}. We evaluated the outcomes under both conditions in two ways: (a) \textit{Answer correctness}: did the model get the question right or wrong? and (b) \textit{Relevance agreement}: how well did the model align with the physician trainee’s annotated relevant components? 

We quantified alignment using metrics such as the proportion of the model’s referenced high/low relevance components and the frequency of referencing irrelevant components, comparing these against the ground-truth annotations. Our goal is to ensure that relevance agreement aligns with both accurate prediction (true positives (TP) in Table \ref{Confusion_Matrix}) and correct relevance, using clinicians' relevance labels with correct predictions as ground truth. We seek to minimize cases where the model achieves correct predictions but relies on incorrect relevance (false positives (FP) in Table \ref{Confusion_Matrix}).

\begin{table}[h]
\centering
\footnotesize  % <-- Make the table font smaller
\renewcommand{\arraystretch}{1.2}  % Slightly reduce row height
\begin{tabular}{@{}llcc@{}}
\toprule
 & & \multicolumn{2}{c}{\textbf{Relevance Agreement}} \\
 & & \textbf{Yes} & \textbf{No} \\
\midrule
\multirow{2}{*}{\textbf{Accurate Prediction}} 
& \textbf{Yes}  
& \cellcolor{green!25}TP (Relevance\checkmark, Accurate \checkmark) 
& \cellcolor{red!40}FP (Relevance\ding{55}, Accurate \checkmark) \\
& \textbf{No}  
& \cellcolor{red!40}FN (Relevance\checkmark, Accurate \ding{55}) 
& \cellcolor{red!25}TN (Relevance\ding{55}, Accurate \ding{55}) \\
\bottomrule
\end{tabular}
\caption{\textbf{Confusion matrix of prediction accuracy and relevance agreement.} In our MedPAIR benchmark, we evaluated relevance labels from both physician trainee labelers and LLMs.}
\label{Confusion_Matrix}
\end{table}

\section{Labeler Instructions \& Prompts}

We asked each physician trainee labeler to follow this instruction during sentence-level relevance labeling:

\begin{tcolorbox}[palebluebox]
\small
You are given a list of sentences from a clinical vignette and a multiple-choice clinical question.

Your task is twofold:
(1) Select the most appropriate answer from the given options.
(2) Label each sentence as either \textbf{[High Relevance]}, \textbf{[Low Relevance]}, or \textbf{[Irrelevant]}, based on its contribution to answering the question.

\vspace{1mm}
\textbf{\textsc{Definitions:}} 

\vspace{2mm}
\textbf{\textsc{[High Relevance]: }} Give this label to sentences that directly answer the medical question with specific and essential information. If this part is missing or altered, the answer would be significantly affected.
\begin{itemize}
    \item A sentence that explicitly states the primary cause or contributing factor (history, demographics, etc.)  is considered high relevance. 
    \item If the question is asking about the treatment plan, a sentence that clearly states the specific indication of the proposed treatment plan is considered high relevance.
    \item If the question is asking about the diagnosis, a sentence that includes diagnostic criteria for the condition is considered high relevance.
    \item If the question is asking about test results, a sentence that clearly reports the key findings that confirm or support the test outcome is considered high relevance.
\end{itemize}

\vspace{2mm}
\textbf{\textsc{[Low Relevance]: }}Give this label to sentences that offer background or contextually related or background information that may be helpful but do not directly answer the question.
\begin{itemize}
    \item A sentence that includes a secondary or potential contributing factor (symptoms, history, etc.) of the main patient condition is considered low relevance. 
    \item A negative history that contradicts or does not support the diagnosis (e.g., no prior epistaxis when the diagnosis is epistaxis) is considered low relevance.
    \item If the question is asking about the treatment plan, a sentence that includes the intervention or therapy that is not central to the gold standard treatment is considered low relevance.
    \item If the question is asking about the treatment plan, a sentence that describes the outcome of a previous intervention for the current chief complaint or diagnosis—particularly one that was unsuccessful—is considered low relevance.
    \item If the question is asking about the treatment plan, a sentence that does not indicate the treatment itself but instead rules out other conditions that would require different treatments is considered low relevance.
    \item If the question is asking about the diagnosis, a sentence that includes criteria that could rule out the current diagnosis (that provides differential diagnosis of the patient condition) is considered low relevance.
    \item If the question is asking about test results, a sentence that reports findings that correlate with or commonly co-occur with the expected result—but are not definitive—is considered low relevance.

\end{itemize}

\vspace{2mm}
\textbf{\textsc{[Irrelevant]: }}Give this label to sentences that do not fall under high or low relevance, or that seem completely unrelated or unhelpful to answering the question. Irrelevant sentences wouldn’t affect anyone answering this QA even if this is removed.
\begin{itemize}
    \item Sentence that adds no additional information on solving question and doesn’t help in differentially diagnosing the condition
    \item General findings, not specific to the diagnosis or management decision.
\end{itemize}

\vspace{2mm}
Focus on identifying the information that directly contributes to answering the question. This task involves only text and does not include any images. If the text refers to figures or mentions ‘from the image,’ focus only on the information presented in the text.
Please consider the following clinical question and answer options when labeling each sentence. Then, label each sentence.
\label{LLM_Self_Report_Prompt}
\end{tcolorbox}

To ensure both physician labelers and the LLM received identical instructions, we used the same prompt when eliciting self-reported sentence-level relevance annotations.

We also asked LLMs to output the answer while compiling the ContextCite score for each sentence.

\begin{tcolorbox}[palebluebox]
\small
You are a clinical reasoning assistant. You will receive a patient case summary and a multiple-choice question. 

Read the question and state your answer. 

Patient Context: [\textit{patient profile text}]

Question and Options: [\textit{question and options}]

\vspace{2mm}

Please select the single most appropriate answer. Respond only in the following format: 

Answer: <LETTER>
\label{Context_Cite_Prompt}
\end{tcolorbox}

\begin{figure}[h!]
  \centering
  \includegraphics[width=\textwidth]{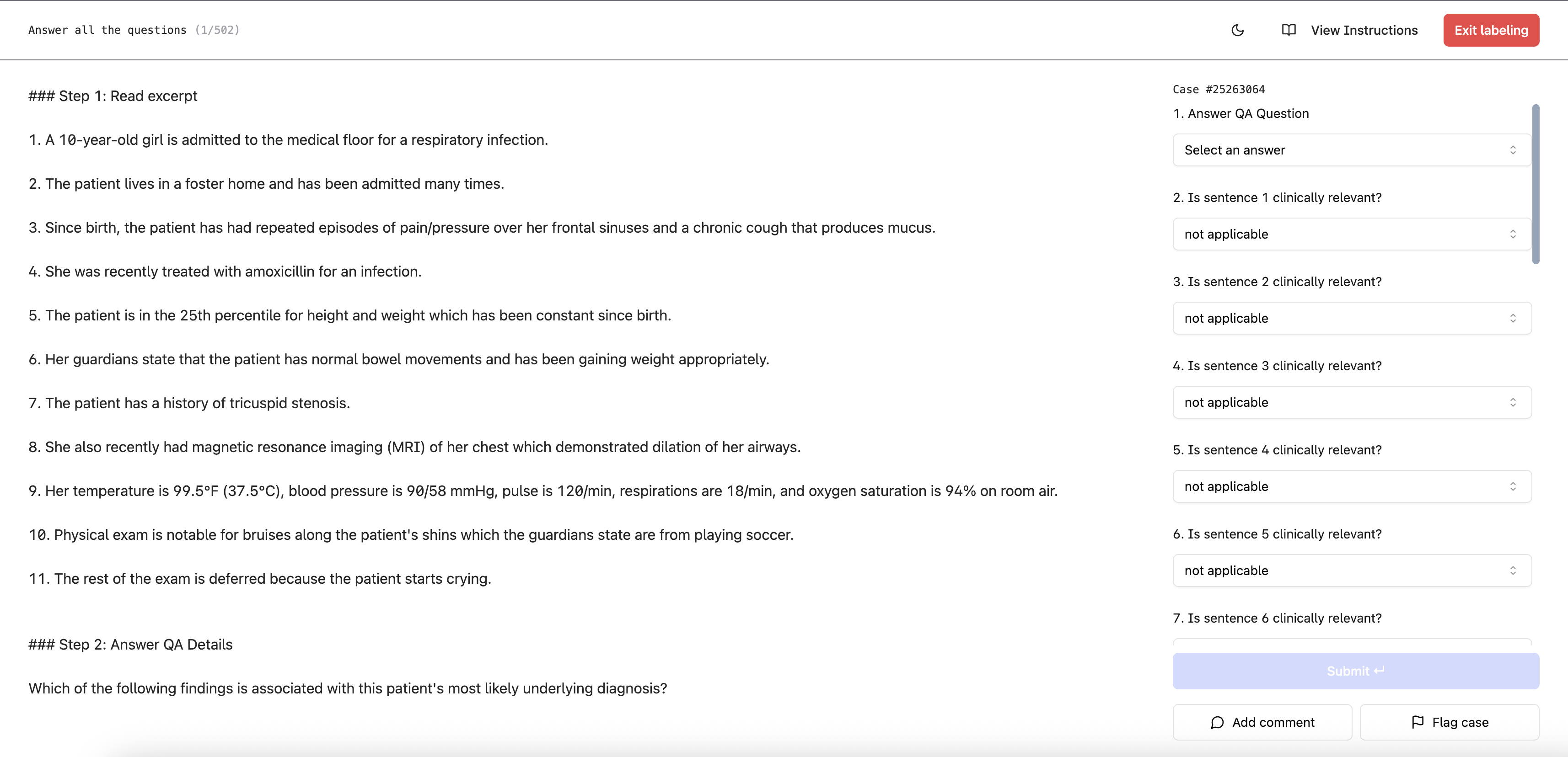}
  \caption{\textbf{Centaur Labs Labeling Interface.} The physician trainee labelers first answer the classification question, then provide high relevance, low relevant, and not relevant labels to each sentence.}
  \label{Interface}
\end{figure}

\begin{table}[h!]
\centering
\resizebox{\textwidth}{!}{%
\begin{tabular}{l|cc|cc|cc|cc|cc}
\toprule
& \multicolumn{2}{c|}{\makecell[c]{MMLU \\ Precision Medicine \\ (193 QAs)}} 
& \multicolumn{2}{c|}{\makecell[c]{JAMA \\ Clinical Challenge \\ (582 QAs)}} 
& \multicolumn{2}{c|}{\makecell[c]{Med Bullets \\ (207 QAs)}} 
& \multicolumn{2}{c|}{\makecell[c]{MedXpertQA \\ (318 QAs)}} 
& \multicolumn{2}{c}{\textbf{\makecell[c]{Overall \\ (1,300 QAs)}}} \\
\midrule
& Before & After & Before & After & Before & After & Before & After & Before & After \\
\midrule
\makecell[l]{Physician \\ Trainees}     & 84.2 {\scriptsize(0.4)} & \makecell[c]{[31 QAs] \\89.3 {\scriptsize(0.2)}} & 55.2 {\scriptsize(0.5)} & \makecell[c]{[93 QAs] \\64.5 {\scriptsize(0.2)}} & 65.8 {\scriptsize(0.5)} & \makecell[c]{[31 QAs] \\76.4 {\scriptsize(0.2)}} & \textbf{23.5} {\scriptsize(0.4)} & \makecell[c]{[93 QAs] \\59.5 {\scriptsize(0.2)}} & \textbf{48.3} {\scriptsize(0.5)} & \makecell[c]{[248 QAs] \\67.2 {\scriptsize(0.2)}} \\
\midrule
Qwen-14B     & 69.8 {\scriptsize(0.5)} & \textcolor{red}{66.3} {\scriptsize(0.5)} & 51.4 {\scriptsize(0.5)} & 52.8 {\scriptsize(0.5)} & 40.9 {\scriptsize(0.5)} & 44.2 {\scriptsize(0.5)} & 18.9  {\scriptsize(0.4)} & 21.1 {\scriptsize(0.4)} & 45.0 {\scriptsize(0.5)} & 45.7 {\scriptsize(0.5)}\\
LLaMA-70B    & 77.9 {\scriptsize(0.42)} & 92.2 {\scriptsize(0.3)} & 48.1 {\scriptsize(0.5)} & 66.8 {\scriptsize(0.47)} & 48.7 {\scriptsize(0.5)} & 69.6 {\scriptsize(0.5)} & 14.9 {\scriptsize(0.4)} & 21.4 {\scriptsize(0.4)} & 30.0 {\scriptsize(0.5)} & 62.2 {\scriptsize(0.5)}\\
Qwen-72B    & 89.3 {\scriptsize(0.3)} & 92.2 {\scriptsize(0.3)} & 57.9 {\scriptsize(0.5)} & 67.9 {\scriptsize(0.5)} & 62.4 {\scriptsize(0.49)} & 67.6 {\scriptsize(0.5)} & 16.2 {\scriptsize(0.4)} & 27.0 {\scriptsize(0.4)} & 35.1 {\scriptsize(0.5)} & 61.5 {\scriptsize(0.5)}\\
GPT-4o     & \textbf{95.6} {\scriptsize(0.2)} & 96.4 {\scriptsize(0.2)}  & \textbf{68.5} {\scriptsize(0.5)} & 78.7 {\scriptsize(0.4)} & \textbf{74.5} {\scriptsize(0.44)} & 84.1 {\scriptsize(0.4)} & 16.4 {\scriptsize(0.4)} & 41.2 {\scriptsize(0.5)}  & 39.3 {\scriptsize(0.3)} & 73.0 {\scriptsize(0.2)} \\
\bottomrule
\end{tabular}
}
\vspace{0.7mm}

\caption{\textbf{Comparison of accuracy (\%) across datasets before and after removing sentences that physician trainees labeled as low relevance or irrelevant.} Values in parentheses represent the corresponding standard deviations. \textbf{Bold} denotes the best performance across all physician trainee labelers and LLMs.  
\textcolor{red}{Red highlighting} denotes a drop relative to the baseline.}
\label{tab:comparison}
\end{table}

\begin{table}[b!]
\centering
\small
\begin{tabular}{l|c|c|c|c}
\toprule
\textbf{Dataset} & \multicolumn{2}{c}{\textbf{Readability}} & \multicolumn{2}{c}{\textbf{Top Keywords Frequency}} \\\hline
& \textbf{High} & \textbf{Low/Irr} & \textbf{High} & \textbf{Low/Irr} \\
\toprule
\midrule
\textbf{\makecell[l]{MMLU \\(Precision \\ Medicine)}}     & 62.0 {\scriptsize(15.3)} & 67.4 {\scriptsize(26.3)} & \makecell[c]{\textcolor{violet}{days}, emergency,\\ department, \textcolor{red}{shortness} \\ \textcolor{orange}{leukocyte, urine}, \\ \textcolor{violet}{previously}} & \makecell[c]{\textcolor{gray}{unremarkable}, \\ \textcolor{violet}{currently}, \textcolor{brown}{smoke}, \textcolor{gray}{controlled}, \\ \textcolor{brown}{illicit, bmi}, illness, \\oral, kgm, weighs}\\\hline
\textbf{\makecell[l]{JAMA \\ Clinical \\ Challenge}}    & 34.5 {\scriptsize(15.3)} & 40.1 {\scriptsize(18.3)} & \makecell[c]{\textcolor{green}{foveal}, \\hyperreflective,\\ spots, \textcolor{brown}{girl}, \\ \textcolor{violet}{progressively}, \textcolor{blue}{hypopyon}, \\ \textcolor{green}{cytoplasm}, punctate, \\  \textcolor{green}{ventricular}, \textcolor{brown}{man}} & \makecell[c]{\textcolor{orange}{swab}, \textcolor{brown}{travel}, \textcolor{orange}{procedures}, \\ \textcolor{orange}{order}, \textcolor{brown}{allergic}, \\ \textcolor{gray}{noncontributory}, pertinent, \\ \textcolor{green}{digital}, empirically, \\\textcolor{brown}{animals}}\\\hline
\textbf{MedBullets}    & 67.0 {\scriptsize(12.2)} & 71.9 {\scriptsize(16.4)} & \makecell[c]{\textcolor{green}{extremity}, \textcolor{violet}{increased},\\ meq/L, \textcolor{violet}{developed}, \\flexion, bright, \\\textcolor{green}{right, lateral}, \\poor, \textcolor{violet}{progressively}} & \makecell[c]{\textcolor{teal}{metformin}, sexually,\\ active, \textcolor{blue}{clear}, \textcolor{teal}{medications}, \\ \textcolor{gray}{uncomplicated, known}, \\ \textcolor{green}{cervical}, \textcolor{blue}{nonfocal}, \textcolor{teal}{albuterol}}\\\hline
\textbf{MedXpertQA}     & 49.8 {\scriptsize(16.0)} & 54.9 {\scriptsize(21.5)} & \makecell[c]{\textcolor{violet}{progressive}, \textcolor{red}{severe},\\ levels, \textcolor{blue}{low}, \textcolor{orange}{urea}, \textcolor{green}{spine}, \\ \textcolor{orange}{nitrogen}, labor, \\ \textcolor{green}{iliac}, mmoll} & \makecell[c]{trauma, \textcolor{brown}{allergies},\\ appropriately, \textcolor{blue}{murmurs}, \\ \textcolor{brown}{vitamin, beers}, \\ personal, \textcolor{orange}{resuscitation}, \\ ordered, \textcolor{brown}{taking}}\\
\bottomrule
\textbf{Overall}     & 47.5 {\scriptsize(19.9)} & 52.9 {\scriptsize(23.9)} & \makecell[c]{spots, \textcolor{green}{foveal}, \\ \textcolor{blue}{hyperreflective}, \textcolor{violet}{progressive}, \\ \textcolor{blue}{hypopyon}, \textcolor{blue}{watery}, cytoplasm, \\punctate, particularly, \textcolor{green}{wrist}} & \makecell[c]{\textcolor{blue}{organomegaly}, \\\textcolor{brown}{smoke, walks}, \\ \textcolor{gray}{comfortable}, \textcolor{gray}{noncontributory}, \\ personal, \textcolor{blue}{nonfocal}, \\ \textcolor{teal}{antihypertensive}, \textcolor{violet}{weekends}, case}\\
\bottomrule
\end{tabular}
\vspace{0.7mm}
\caption{\textbf{Comparison of dataset characteristics focusing on Readability and Top Keywords Frequency.} Values in parentheses represent standard deviations. The readability is calculated through Flesch Reading Ease score, which typically ranges from 0 to 100, where a higher score indicates that the text is easier to read, and a lower score suggests the text is more difficult. We highlighted each clinical term using different colors based on the type of information it conveys: \textcolor{red}{symptoms, severity}, \textcolor{blue}{description on findings}, \textcolor{brown}{demographics / history}, \textcolor{teal}{medicine}, \textcolor{orange}{medical test}, \textcolor{green}{anatomical structure/term}, \textcolor{gray}{negative findings or suggestive of good patient status}, \textcolor{violet}{timeline, comparative}.}
\label{tab:keyword_readability}
\end{table}

\section{Sample QA Case Study}
\label{sample}
\begin{tcolorbox}[colback=blue!3!white, colframe=blue!50!white,
                  title=Example of QA Data, left=10mm, breakable]
\small

\textbf{Patient Profile:}
\begin{enumerate}[leftmargin=*]
  \item A 29-year-old female presents with low back pain of five days’ duration.
  \item Her new job involves walking several miles daily across a large facility.
  \item The pain is localized without radiation; no traumatic history.
  \item Medications: only oral contraceptives.
\end{enumerate}

\bigskip
\textbf{Question: }What is the most likely diagnosis?

\bigskip
\textbf{Options:}
\begin{enumerate}[label=\Alph*.]
  \item bilateral sacral extension  
  \item bilateral sacral flexion  
  \item sacral base posterior  
  \item right-on-right sacral torsion  
  \item sacral base anterior  
  \item right-on-left sacral torsion  
  \item unilateral sacral flexion on the right  
  \item left-on-left sacral torsion  
  \item left-on-right sacral torsion  
  \item unilateral sacral extension on the left  
\end{enumerate}

\bigskip
\textbf{Correct Answer:} (D) right-on-right sacral torsion.

\bigskip
\noindent
\label{tab:label_landscape_comparison}

\medskip
\centering
\scriptsize
\setlength{\tabcolsep}{4pt}
\begin{tabular}{|l|l|l|l|}
  \toprule
  Sentence \# & Physician Labels & GPT4o Self-Reported Labels & Llama70B ContextCite Labels \\
  \midrule
  1 & High & High & High\\
  2 & Low  & Low  & High\\
  3 & Low  & High & High\\
  4 & Irr  & Irr  & Low \\   
  \bottomrule         
\end{tabular}
\end{tcolorbox}

In this case study, we observe notable disagreement in informativeness assessments across sentence 2 and sentence 3 among physicians, GPT-4o, and LLaMA-70B ContextCite. Sentence 2 (“Her new job involves walking several miles daily across a large facility”) was labeled as \textit{Low} by both physicians and GPT-4o, yet \textit{High} by LLaMA-70B ContextCite. This sentence describes the patient’s lifestyle, specifically her physical activity level related to her job. This information is of limited relevance to sacral torsion. While one cannot entirely rule out its contribution—since prolonged walking with an asymmetric posture could potentially predispose a patient to sacral dysfunction—it is not a direct cause or a diagnostically decisive factor. As such, it offers minimal value in determining the correct answer to this question. LLaMA-70B may have overemphasized contextual lifestyle clues, interpreting the exertion from walking as highly indicative of a mechanical sacral dysfunction, whereas clinicians likely viewed it as a nonspecific background factor without clear diagnostic utility. 

The sentence 3 (“The pain is localized without radiation; no traumatic history”) received a Low label from physicians but High from GPT-4o and LLaMA-70B. This discrepancy may reflect differing heuristics: while clinicians might not prioritize localization and absence of trauma due to their non-specificity or commonality in musculoskeletal complaints, models may have heuristically linked "localized pain without radiation" to mechanical causes, interpreting it as informative. These examples illustrate how LLMs may misattribute diagnostic weight to surface-level patterns.

\section{Qualitative Survey Analysis}
\subsection{Pre-Study Survey}
In the pre‑study survey all 36 labelers provided demographic and background information. The cohort was predominantly male (69.4\%), with female annotators accounting for 27.8\%. Most participants (85.7\%) were in the senior years of their medical training, and 72.2\% had already passed the USMLE Step 1, underscoring their competence in tackling medical QA tasks. Rare‑case resources were well known to the group: 52.8\% reported familiarity with collections such as the JAMA Clinical Challenge, NEJM Image Challenge, and NEJM Resident 360, and 41.7\% stated that they consult these materials regularly. Exposure to clinical large language models was also high: 91.7\% had used or observed LLMs in practice, whether for answering clinical questions or integrating decision‑support functions. On average, respondents estimated that 49.0\% (SD = 21.8) of existing LLMs are sufficiently mature for deployment in clinical settings.

\subsection{Post-Study Survey}
We received 29 post-study surveys from the 36 participating labelers. On average, only 8.6\% (SD = 0.2) of the questions they labeled had been encountered verbatim before the study. Labelers expected large language models to outperform them, predicting 79.7\% accuracy for the models (SD = 0.1) versus 63.5\% for themselves (SD= 0.2). When reflecting on their sentence‑level relevance judgments, a majority of 72.4\% described themselves as “moderately confident,” an 58.6\% characterized the task as exhibiting “moderate” ambiguity. Regarding how well the multiple choice QA format matches real-world clinical practice, 44. 8\% viewed the multiple choice QA format as 'moderately aligned' with clinical practice, while 37. 9\% considered it 'slightly aligned. As labelers are allowed to skip questions if they do not know the answer, an average of 20.9\% (12.9) of questions are skipped, which demonstrates the difficulties of the QAs. These qualitative feedback suggests that relevance labels represent consensus‑based approximations rather than definitive ground truth, and that QA benchmarks should be complemented by richer, practice‑grounded evaluations in future work.

Both the pre-study and post-study survey results are available in the supplementary material. 

% \newpage
% \section{Physician Trainee Labeler Information}
\begin{landscape}
\begin{table}[]
\scriptsize
\begin{tabular}{rllllllllll}

\toprule
\multicolumn{1}{c}{\textbf{Age}} & \multicolumn{1}{c}{\textbf{Gender}} & \multicolumn{1}{c}{\textbf{\makecell[l]{Year of \\med school?}}} & \multicolumn{1}{c}{\textbf{\makecell[l]{USMLE \\Step 1 \\passed?}}} & \multicolumn{1}{c}{\textbf{\makecell[l]{Medical school}}} & \multicolumn{1}{c}{\textbf{\makecell[l]{Familiarity with \\clinical challenges? \\ (i.e. JAMA \\Clinical Challenge, \\NEJM Image \\Challenge, NEJM \\Resident 360.)}}} & \multicolumn{1}{c}{\textbf{\makecell[l]{If you are familiar \\with any of the \\clinical challenges, how\\ regularly \\do you follow these \\challenges?}}} & \multicolumn{1}{c}{\textbf{\makecell[l]{Familiarity with\\ MedBullets}}} & \multicolumn{1}{c}{\textbf{\makecell[l]{How often do you\\ follow clinical challenges \\such as JAMA/NEJM\\ Challenge?}}} & \multicolumn{1}{c}{\textbf{\makecell[l]{Familiarity with LLMs\\ in healthcare}}} & \multicolumn{1}{c}{\textbf{\makecell[l]{Percentage (\%) \\LLM Clinical \\deployment \\readiness percentage}}} \\
\midrule

N/A & Female & M3 & No & Case Western & Some familiarity & Not at all & High familiarity & Not at all & High familiarity & 30 \\\hline
28 & Male & \makecell[l]{G3 \\(MD/PhD)} & Yes & UC San Diego & Some familiarity & Not at all & Not familiar & N/A & High familiarity & 30 \\\hline
24 & Male & M3 & Yes & Columbia VP\&S & Not familiar & N/A & Not familiar & N/A & High familiarity & 20 \\\hline
N/A & Male & M2 & Yes & \makecell[l]{NYU Grossman School\\ of Medicine} & Some familiarity & Not at all & Not familiar & Not at all & Some familiarity & 70 \\\hline
25 & Male & M3 & Yes & UNC School of Medicine & Not familiar & N/A & Not familiar & N/A & High familiarity & 80 \\\hline
26 & Male & M4 & Yes & Dell Medical school & Some familiarity & Not at all & Not familiar & Not at all & Some familiarity & 30 \\\hline
27 & Queer & M3 & Yes & KPSOM & Some familiarity & Not at all & Some familiarity & Not at all & Not familiar & \multicolumn{1}{l}{N/A} \\\hline
25 & Male & M4 & Yes & University of Toledo & Some familiarity & Not at all & High familiarity & Not at all & Some familiarity & 25 \\\hline
N/A & Male & M2 & Yes & Harvard & High familiarity & Occasionally & High familiarity & Occasionally & High familiarity & 25 \\\hline
26 & Female & M4 & Yes & \makecell[l]{George Washington \\University SOM} & High familiarity & Occasionally & High familiarity & Occasionally & High familiarity & 90 \\\hline
27 & Female & M3 & Yes & UNC Chapel Hill SOM & Not familiar & N/A & Some familiarity & Not at all & Some familiarity & 30 \\\hline
26 & Male & M4 & Yes & UNC Chapel Hill & Not familiar & Not at all & Not familiar & Not at all & Some familiarity & 30 \\\hline
28 & Female & M3 & Yes & KPSOM & Some familiarity & Occasionally & High familiarity & Occasionally & High familiarity & 40 \\\hline
25 & Male & M3 & No & WUSM & Not familiar & Not at all & Some familiarity & Not at all & High familiarity & 70 \\\hline
26 & Female & M3 & Yes & \makecell[l]{Tufts University \\School of Medicine} & Not familiar & Not at all & Not familiar & Not at all & Some familiarity & 40 \\\hline
26 & Female & M3 & No & \makecell[l]{Tufts University\\ School of Medicine} & Not familiar & Not at all & Not familiar & N/A & Not familiar & 70 \\\hline
22 & Male & M1 & No & \makecell[l]{Dartmouth Geisel \\School of Medicine} & Some familiarity & Not at all & High familiarity & Occasionally & High familiarity & 80 \\\hline
25 & Male & M4 & Yes & University of Toledo & Some familiarity & Occasionally & Some familiarity & Occasionally & High familiarity & 80 \\\hline
26 & Female & M4 & Yes & Medical College of Georgia & Not familiar & N/A & Some familiarity & Not at all & Some familiarity & 45 \\\hline
28 & Male & M4 & Yes & \makecell[l]{Alabama College \\of Osteopathic Medicine} & Some familiarity & Occasionally & Not familiar & Not at all & Some familiarity & 60 \\ \hline
25 & Male & M3 & Yes & \makecell[l]{Warren Alpert Medical School \\of Brown University} & Some familiarity & Occasionally & High familiarity & Not at all & Some familiarity & 45 \\ \hline
32 & Male & M4 & Yes & Northwestern & Not familiar & Not at all & Some familiarity & Not at all & Some familiarity & 5 \\ \hline
27 & Female & M4 & Yes & \makecell[l]{Alabama College \\of Osteopathic Medicine} & Not familiar & N/A & Not familiar & N/A & Some familiarity & 50 \\ \hline
23 & Male & M1 & No & \makecell[l]{University of Maryland \\School of medicine} & Some familiarity & Occasionally & Some familiarity & Not at all & Some familiarity & 45 \\\hline
N/A & Male & M4 & Yes & Harvard & High familiarity & Occasionally & Not familiar & Not at all & High familiarity & 60 \\\hline
28 & Female & M4 & Yes & Dartmouth & Not familiar & N/A & Some familiarity & Not at all & High familiarity & 80 \\\hline
25 & Male & M3 & Yes & \makecell[l]{Indiana University\\ School of Medicine} & Some familiarity & Not at all & Some familiarity & Not at all & High familiarity & 20 \\\hline
29 & Male & M4 & Yes & Emory University & Some familiarity & Occasionally & Not familiar & Not at all & High familiarity & 60 \\\hline
23 & Male & M1 & No & UC Irvine & Not familiar & Not at all & Not familiar & Not at all & High familiarity & 50 \\\hline
33 & Female & M4 & Yes & \makecell[l]{Touro College of Osteopathic \\Medicine Middletown NY} & Not familiar & N/A & Not familiar & N/A & Some familiarity & 70 \\\hline
26 & Male & M4 & Yes & UNC School of Medicine & Some familiarity & Occasionally & High familiarity & Occasionally & High familiarity & 75 \\\hline
25 & Male & M1 & No & \makecell[l]{University of Texas\\ Medical Branch} & Some familiarity & Occasionally & Not familiar & Not at all & Some familiarity & 50 \\\hline
N/A & Male & M4 & Yes & Dell Medical School & Not familiar & N/A & Some familiarity & Not at all & Not familiar & 20 \\\hline
26 & Male & M3 & No & Rush Medical College & Some familiarity & Occasionally & Some familiarity & Occasionally & High familiarity & 25 \\
\bottomrule
\end{tabular}
\vspace{2mm}
\caption{\textbf{Pre-Survey Demographics and Educational Background of Medical Student Participants, Including Self-Reported Familiarity with Clinical Challenges and LLMs in Healthcare.} "N/A" denotes that the labeler chose not to disclose this information. The complete list of pre-survey questions is available in the supplementary material.}
\label{demographic}
\end{table}
\end{landscape}

\newpage
\newpage

\section*{NeurIPS Paper Checklist}
\begin{enumerate}

\item {\bf Claims}
    \item[] Question: Do the main claims made in the abstract and introduction accurately reflect the paper's contributions and scope?
    \item[] Answer: \textcolor{blue}{Yes}
    \item[] Justification: \textcolor{blue}{Contributions are clearly enumerated at the end of the introduction, highlighting results and resources that can be found within the manuscript.} 
    % \item[] Guidelines:
    % \begin{itemize}
    %     \item The answer NA means that the abstract and introduction do not include the claims made in the paper.
    %     \item The abstract and/or introduction should clearly state the claims made, including the contributions made in the paper and important assumptions and limitations. A No or NA answer to this question will not be perceived well by the reviewers. 
    %     \item The claims made should match theoretical and experimental results, and reflect how much the results can be expected to generalize to other settings. 
    %     \item It is fine to include aspirational goals as motivation as long as it is clear that these goals are not attained by the paper. 
    % \end{itemize}

\item {\bf Limitations}
    \item[] Question: Does the paper discuss the limitations of the work performed by the authors?
    \item[] Answer: \textcolor{blue}{Yes}
    \item[] Justification: \textcolor{blue}{A dedicated limitations section can be found at the end of the paper. This highlights key methodological limitations and explains attempts to address robustness, particularly with respect to template variation.} 

\item {\bf Theory Assumptions and Proofs}
    \item[] Question: For each theoretical result, does the paper provide the full set of assumptions and a complete (and correct) proof?
    \item[] Answer: \textcolor{blue}{N/A}.
    \item[] Justification: \textcolor{blue}{No theoretical results are presented in this piece. Any calculations have associated equations in-line and are referenced as such.}
    % \item[] Guidelines:
    % \begin{itemize}
    %     \item The answer NA means that the paper does not include theoretical results. 
    %     \item All the theorems, formulas, and proofs in the paper should be numbered and cross-referenced.
    %     \item All assumptions should be clearly stated or referenced in the statement of any theorems.
    %     \item The proofs can either appear in the main paper or the supplemental material, but if they appear in the supplemental material, the authors are encouraged to provide a short proof sketch to provide intuition. 
    %     \item Inversely, any informal proof provided in the core of the paper should be complemented by formal proofs provided in appendix or supplemental material.
    %     \item Theorems and Lemmas that the proof relies upon should be properly referenced. 
    % \end{itemize}

    \item {\bf Experimental Result Reproducibility}
    \item[] Question: Does the paper fully disclose all the information needed to reproduce the main experimental results of the paper to the extent that it affects the main claims and/or conclusions of the paper (regardless of whether the code and data are provided or not)?
    \item[] Answer: \textcolor{blue}{Yes}
    \item[] Justification: \textcolor{blue}{All code is available in a public repository that enables the running the context reduction guided by physician annotations and ContextCite scores from open-source LLMs.} 

\item {\bf Open access to data and code}
    \item[] Question: Does the paper provide open access to the data and code, with sufficient instructions to faithfully reproduce the main experimental results, as described in supplemental material?
    \item[] Answer: \textcolor{blue}{Yes}
    \item[] Justification: \textcolor{blue}{A detailed README has been provided within each repository folder describing the steps required to reproduce or extend the current work. }
    % \item[] Guidelines:
    % \begin{itemize}
    %     \item The answer NA means that paper does not include experiments requiring code.
    %     \item Please see the NeurIPS code and data submission guidelines (\url{https://nips.cc/public/guides/CodeSubmissionPolicy}) for more details.
    %     \item While we encourage the release of code and data, we understand that this might not be possible, so “No” is an acceptable answer. Papers cannot be rejected simply for not including code, unless this is central to the contribution (e.g., for a new open-source benchmark).
    %     \item The instructions should contain the exact command and environment needed to run to reproduce the results. See the NeurIPS code and data submission guidelines (\url{https://nips.cc/public/guides/CodeSubmissionPolicy}) for more details.
    %     \item The authors should provide instructions on data access and preparation, including how to access the raw data, preprocessed data, intermediate data, and generated data, etc.
    %     \item The authors should provide scripts to reproduce all experimental results for the new proposed method and baselines. If only a subset of experiments are reproducible, they should state which ones are omitted from the script and why.
    %     \item At submission time, to preserve anonymity, the authors should release anonymized versions (if applicable).
    %     \item Providing as much information as possible in supplemental material (appended to the paper) is recommended, but including URLs to data and code is permitted.
    % \end{itemize}

\item {\bf Experimental Setting/Details}
    \item[] Question: Does the paper specify all the training and test details (e.g., data splits, hyperparameters, how they were chosen, type of optimizer, etc.) necessary to understand the results?
    \item[] Answer:\textcolor{blue}{Yes}
    \item[] Justification: \textcolor{blue}{No training or tuning was conducted.}
    % \item[] Guidelines:
    % \begin{itemize}
    %     \item The answer NA means that the paper does not include experiments.
    %     \item The experimental setting should be presented in the core of the paper to a level of detail that is necessary to appreciate the results and make sense of them.
    %     \item The full details can be provided either with the code, in appendix, or as supplemental material.
    % \end{itemize}

\item {\bf Experiment Statistical Significance}
    \item[] Question: Does the paper report error bars suitably and correctly defined or other appropriate information about the statistical significance of the experiments?
    \item[] Answer: \textcolor{blue}{Yes}.
    \item[] Justification: \textcolor{blue}{The paper reports the statistical significance of the experiments, providing error bars and appropriate analysis of performance across multiple methods. The use of statistical measures enhances the credibility of the findings, demonstrating that the improvements in performance are meaningful and not due to random chance.}
    % \item[] Guidelines:
    % \begin{itemize}
    %     \item The answer NA means that the paper does not include experiments.
    %     \item The authors should answer "Yes" if the results are accompanied by error bars, confidence intervals, or statistical significance tests, at least for the experiments that support the main claims of the paper.
    %     \item The factors of variability that the error bars are capturing should be clearly stated (for example, train/test split, initialization, random drawing of some parameter, or overall run with given experimental conditions).
    %     \item The method for calculating the error bars should be explained (closed form formula, call to a library function, bootstrap, etc.)
    %     \item The assumptions made should be given (e.g., Normally distributed errors).
    %     \item It should be clear whether the error bar is the standard deviation or the standard error of the mean.
    %     \item It is OK to report 1-sigma error bars, but one should state it. The authors should preferably report a 2-sigma error bar than state that they have a 96\% CI, if the hypothesis of Normality of errors is not verified.
    %     \item For asymmetric distributions, the authors should be careful not to show in tables or figures symmetric error bars that would yield results that are out of range (e.g. negative error rates).
    %     \item If error bars are reported in tables or plots, The authors should explain in the text how they were calculated and reference the corresponding figures or tables in the text.
    % \end{itemize}

\item {\bf Experiments Compute Resources}
    \item[] Question: For each experiment, does the paper provide sufficient information on the computer resources (type of compute workers, memory, time of execution) needed to reproduce the experiments?
    \item[] Answer: \textcolor{blue}{Yes}
    \item[] Justification: \textcolor{blue}{We used 2 A100 GPUs for open-source LLM ContextCite Score generation. Other experiments used 2 CPUs or 1 A6000 GPU.}
    % \item[] Guidelines:
    % \begin{itemize}
    %     \item The answer NA means that the paper does not include experiments.
    %     \item The paper should indicate the type of compute workers CPU or GPU, internal cluster, or cloud provider, including relevant memory and storage.
    %     \item The paper should provide the amount of compute required for each of the individual experimental runs as well as estimate the total compute. 
    %     \item The paper should disclose whether the full research project required more compute than the experiments reported in the paper (e.g., preliminary or failed experiments that didn't make it into the paper). 
    % \end{itemize}
    
\item {\bf Code Of Ethics}
    \item[] Question: Does the research conducted in the paper conform, in every respect, with the NeurIPS Code of Ethics \url{https://neurips.cc/public/EthicsGuidelines}?
    \item[] Answer: \textcolor{blue}{Yes}
    \item[] Justification: \textcolor{blue}{This authors of this study have read, and confirm this study conforms with every aspect of the Code of Ethics.}
    % \item[] Guidelines:
    % \begin{itemize}
    %     \item The answer NA means that the authors have not reviewed the NeurIPS Code of Ethics.
    %     \item If the authors answer No, they should explain the special circumstances that require a deviation from the Code of Ethics.
    %     \item The authors should make sure to preserve anonymity (e.g., if there is a special consideration due to laws or regulations in their jurisdiction).
    % \end{itemize}

\item {\bf Broader Impacts}
    \item[] Question: Does the paper discuss both potential positive societal impacts and negative societal impacts of the work performed?
    \item[] Answer: \textcolor{blue}{Yes}
    \item[] Justification: \textcolor{blue}{Our work has no negative socieal impacts.}

\item {\bf Safeguards}
    \item[] Question: Does the paper describe safeguards that have been put in place for responsible release of data or models that have a high risk for misuse (e.g., pretrained language models, image generators, or scraped datasets)?
    \item[] Answer:\textcolor{blue}{N/A}.
    \item[] Justification: \textcolor{blue}{All models and datasets utilized in this study are already publicly available. }
    % \item[] Guidelines:
    % \begin{itemize}
    %     \item The answer NA means that the paper poses no such risks.
    %     \item Released models that have a high risk for misuse or dual-use should be released with necessary safeguards to allow for controlled use of the model, for example by requiring that users adhere to usage guidelines or restrictions to access the model or implementing safety filters. 
    %     \item Datasets that have been scraped from the Internet could pose safety risks. The authors should describe how they avoided releasing unsafe images.
    %     \item We recognize that providing effective safeguards is challenging, and many papers do not require this, but we encourage authors to take this into account and make a best faith effort.
    % \end{itemize}

\item {\bf Licenses for existing assets}
    \item[] Question: Are the creators or original owners of assets (e.g., code, data, models), used in the paper, properly credited and are the license and terms of use explicitly mentioned and properly respected?
    \item[] Answer: \textcolor{blue}{Yes}
    \item[] Justification: \textcolor{blue}{All datasets are open access and comply with the copyright and terms of service.}
    % \item[] Guidelines:
    % \begin{itemize}
    %     \item The answer NA means that the paper does not use existing assets.
    %     \item The authors should cite the original paper that produced the code package or dataset.
    %     \item The authors should state which version of the asset is used and, if possible, include a URL.
    %     \item The name of the license (e.g., CC-BY 4.0) should be included for each asset.
    %     \item For scraped data from a particular source (e.g., website), the copyright and terms of service of that source should be provided.
    %     \item If assets are released, the license, copyright information, and terms of use in the package should be provided. For popular datasets, \url{paperswithcode.com/datasets} has curated licenses for some datasets. Their licensing guide can help determine the license of a dataset.
    %     \item For existing datasets that are re-packaged, both the original license and the license of the derived asset (if it has changed) should be provided.
    %     \item If this information is not available online, the authors are encouraged to reach out to the asset's creators.
    % \end{itemize}

\item {\bf New Assets}
    \item[] Question: Are new assets introduced in the paper well documented and is the documentation provided alongside the assets?
    \item[] Answer: \textcolor{blue}{Yes}
    \item[] Justification: \textcolor{blue}{Details of the datasets, counts, code, and findings are all available on Github/Huggingface. We will also provide a blog on this website with a more user-friendly explanation of the approach and findings. We aim to increase accessibility of the results to a broader audience.}
    % \item[] Guidelines:
    % \begin{itemize}
    %     \item The answer NA means that the paper does not release new assets.
    %     \item Researchers should communicate the details of the dataset/code/model as part of their submissions via structured templates. This includes details about training, license, limitations, etc. 
    %     \item The paper should discuss whether and how consent was obtained from people whose asset is used.
    %     \item At submission time, remember to anonymize your assets (if applicable). You can either create an anonymized URL or include an anonymized zip file.
    % \end{itemize}

\item {\bf Crowdsourcing and Research with Human Subjects}
    \item[] Question: For crowdsourcing experiments and research with human subjects, does the paper include the full text of instructions given to participants and screenshots, if applicable, as well as details about compensation (if any)? 
    \item[] Answer: \textcolor{blue}{Yes}
    \item[] Justification:\textcolor{blue}{The detailed instructions are attached in supplementary material. The compensation details are mentioned in the manuscript.}
    % \item[] Guidelines:
    % \begin{itemize}
    %     \item The answer NA means that the paper does not involve crowdsourcing nor research with human subjects.
    %     \item Including this information in the supplemental material is fine, but if the main contribution of the paper involves human subjects, then as much detail as possible should be included in the main paper. 
    %     \item According to the NeurIPS Code of Ethics, workers involved in data collection, curation, or other labor should be paid at least the minimum wage in the country of the data collector. 
    % \end{itemize}

\item {\bf Institutional Review Board (IRB) Approvals or Equivalent for Research with Human Subjects}
    \item[] Question: Does the paper describe potential risks incurred by study participants, whether such risks were disclosed to the subjects, and whether Institutional Review Board (IRB) approvals (or an equivalent approval/review based on the requirements of your country or institution) were obtained?
    \item[] Answer: \textcolor{blue}{Yes}
    \item[] Justification: \textcolor{blue}{We received an IRB2411001474 exempt for our project titled "Towards Digital Sustainability in Health Care: Developing Digital Health Products through Data-Driven User Insights".}
    % \item[] Guidelines:
    % \begin{itemize}
    %     \item The answer NA means that the paper does not involve crowdsourcing nor research with human subjects.
    %     \item Depending on the country in which research is conducted, IRB approval (or equivalent) may be required for any human subjects research. If you obtained IRB approval, you should clearly state this in the paper. 
    %     \item We recognize that the procedures for this may vary significantly between institutions and locations, and we expect authors to adhere to the NeurIPS Code of Ethics and the guidelines for their institution. 
    %     \item For initial submissions, do not include any information that would break anonymity (if applicable), such as the institution conducting the review.
    % \end{itemize}

\end{enumerate}

\end{document}